\documentclass{article}

\usepackage[sort]{natbib}
\PassOptionsToPackage{numbers, compress}{natbib}
\usepackage[final]{format/neurips_2024}

\usepackage[utf8]{inputenc}
\usepackage[T1]{fontenc}
\usepackage{url}
\usepackage{booktabs}
\usepackage{amsfonts}
\usepackage{nicefrac}
\usepackage{microtype}
\usepackage{xcolor}
\usepackage{tablefootnote}
\usepackage[bbgreekl]{mathbbol}
\usepackage{tikz}

\usepackage{subfiles}
\usepackage[final, colorlinks, citecolor=blue]{hyperref}

\usepackage{amsthm}
\usepackage{amssymb}
\usepackage{mathtools}
\usepackage{dsfont}
\usepackage{cleveref}
\usepackage{enumitem}

\newtheorem{theorem}{Theorem}
\newtheorem{lemma}[theorem]{Lemma}

\newtheorem{claim}[theorem]{Claim}
\theoremstyle{definition}
\newtheorem{definition}[theorem]{Definition}
\theoremstyle{definition}

\theoremstyle{remark}

\theoremstyle{definition}

\DeclareMathOperator*{\argmax}{arg\,max}

\DeclareMathOperator{\sign}{\mathrm{sign}}
\DeclareMathOperator{\corr}{\mathrm{corr}}

\DeclareSymbolFontAlphabet{\mathbb}{AMSb}
\DeclareSymbolFontAlphabet{\mathbbl}{bbold}

\newif\ifdraft
\drafttrue
\newcommand\addauthor[3]{
    \expandafter\providecommand\csname #1\endcsname[1]{\ifdraft{
    	{\color{#3} ({\bf #2}| ##1)}
    }{}}
}

\nocite{*}

\usepackage{algorithm,algorithmic}

\title{Sample-Efficient Agnostic Boosting}

\author{%
  Udaya Ghai\\
  Amazon\\
  \texttt{ughai@amazon.com} \\
  \And
  Karan Singh\\
  Tepper School of Business\\
  Carnegie Mellon University\\
  \texttt{karansingh@cmu.edu}
}

\begin{document}

\maketitle

\begin{abstract}
The theory of boosting provides a computational framework for aggregating approximate weak learning algorithms, which perform marginally better than a random predictor, into an accurate strong learner. In the realizable case, the success of the boosting approach is underscored by a remarkable fact that the resultant sample complexity matches that of a computationally demanding alternative, namely Empirical Risk Minimization (ERM). This in particular implies that the realizable boosting methodology has the potential to offer computational relief without compromising on sample efficiency.

Despite recent progress, in agnostic boosting, where assumptions on the conditional distribution of labels given feature descriptions are absent, ERM outstrips the agnostic boosting methodology in being quadratically more sample efficient than all known agnostic boosting algorithms. In this paper, we make progress on closing this gap, and give a substantially more sample efficient agnostic boosting algorithm than those known, without compromising on the computational (or oracle) complexity. A key feature of our algorithm is that it leverages the ability to reuse samples across multiple rounds of boosting, while guaranteeing a generalization error strictly better than those obtained by blackbox applications of uniform convergence arguments. We also apply our approach to other previously studied learning problems, including boosting for reinforcement learning, and demonstrate improved results.
\end{abstract}

\section{Introduction}
A striking observation in statistical learning is that given a small number of samples it is possible to learn the best classifier from an almost exponentially large class of predictors. In fact, it is possible to do using a conceptually straightforward procedure -- Empirical Risk Minimization (ERM) -- that finds a classifier that is maximally consistent with the collected samples. Substantiating this observation, the fundamental theorem of statistical learning (e.g., \citep{shalev2014understanding}) states that with high probability ERM can guarantee $\varepsilon-$excess population error with respect to the best classifier from a finite, but large hypothesis class $\mathcal{H}$ given merely $m_{\text{agnostic}}\approx (\log |\mathcal{H}|)/{\varepsilon^2}$ identically distributed and independent (IID) samples of pairs of features and labels from the population distribution. Under an additional assumption -- that of {\em realizability} -- guaranteeing that there exists a perfect classifier in the hypothesis class achieving zero error, a yet quadratically smaller number  $m_{\text{realizable}}\approx ({\log |\mathcal{H}|})/{\varepsilon}$ of samples suffice. In the absence of such assumption, the learning problem is said to take place in the {\em agnostic} setting, i.e., under of a lack of belief in the ability of any hypothesis to perfectly fit the observed data.

This ability to generalize to the population distribution and successfully (PAC) learn from an almost exponentially large, and hence expressive, hypothesis class given limited number of examples suggests that the primary bottleneck for efficient learning is computational. Indeed, even with modest sample requirements, finding a maximally consistent hypothesis within an almost exponentially large class via, say, enumeration or global search, is generally computationally intractable. 

It is against this backdrop that the theory of boosting offers a compelling alternative. 
The starting point is the realization that often, both in practice and in theory, it is easy to construct 
simple, yet inaccurate {\em rules-of-thumbs} (or {\em weak learners}) that perform ever so slightly better
than a random classifier. A natural question then arises (paraphrased from \cite{schapire1990strength}'s abstract):
can one convert such mediocre learning rules into one that performs extremely well? Boosting algorithms offer a positive
resolution to this question by providing convenient and computationally efficient reductions that aggregate such weak 
learners into a proficient learner with an arbitrarily high accuracy.

\begin{table}
    \begin{tabular}{@{}p{\textwidth}@{}}
        \centering
        \bgroup
        \def\arraystretch{1.6}
        \begin{tabular}{|c|c|c|}
            \hline
            & Sample Complexity & Oracle Complexity  \\
            \hline
            \cite{kanade2009potential}   &    $(\log |\mathcal{B}|)/\gamma^4\varepsilon^4$ & $1/\gamma^2\varepsilon^2$
            \\
            \hline
            \cite{brukhim2020online}   & $(\log |\mathcal{B}|)/\gamma^4\varepsilon^6$ & $1/\gamma^2\varepsilon^2$
            \\
            \hline
            \Cref{thm:maincons} & $(\log |\mathcal{B}|)/\gamma^3\varepsilon^3$ & $(\log |\mathcal{B}|)/\gamma^2\varepsilon^2$\\
            \hline
            \Cref{thm:maincons2} (in \Cref{app:improv}) & $(\log |\mathcal{B}|)/\gamma^3\varepsilon^3 + (\log |\mathcal{B}|)^3/\gamma^2\varepsilon^2$ & $1/\gamma^2\varepsilon^2$\\
            \hline
            ERM (no boosting) & $(\log |\mathcal{B}|)/\gamma^2\varepsilon^2$ & $+\infty$ (inefficient) \\
            \hline
        \end{tabular}
        \egroup
    \end{tabular}
    \caption{A comparison between sample and oracle complexities (i.e., number of weak learning calls) of the present results and previous works, in each case to achieve $\varepsilon$-excess population error. Here we suppress polylogarithmic factors. We make progress on closing the sample complexity gap between ERM, which is computationally inefficient, and boosting-based approaches. The $\gamma$-weak leaner outputs a hypothesis from the base class $\mathcal{B}$, which is usually substantially smaller than $\mathcal{H}$ against which the final agnostic learning guarantee holds. In practice, boosting is used with learners with small values of $\log |\mathcal{B}|$. See \Cref{def:wl} for details. See the paragraph following Theorem 1 in \cite{kanade2009potential} and Section 3.3 in \cite{brukhim2020online} for derivation of these bounds. See also Theorem 2.14 in \cite{alon2021boosting} for a bound on the expressivity of the boosted class to derive ERM's sample complexity.} 
    \label{tab:table1}
\end{table}

\paragraph{Realizable Boosting.} Consider the celebrated Adaboost algorithm \citep{freund1997decision} which operates in the (noiseless) realizable binary classification setting. On any distribution consistent with a fixed labeling function (or {\em concept}), a weak learner here promises an accuracy strictly better than half, since guessing labels randomly gets any example correct with probability half. Given access to such a weak learner and $m_{\text{realizable boosting}}\approx (\log |\mathcal{H}|)/\varepsilon$ samples\footnote{In the introduction, for simplicity, we suppress polynomial dependencies in the weak learner's edge $\gamma$, and polylogarithmic terms. A recent sample complexity lower bound due to \cite{green2022optimal} implies that this equivalence continues to hold even taking into account $\text{poly}(\gamma)$ dependencies as long as $\gamma$ is not exponentially small.}, Adaboost makes $\approx \log 1/\varepsilon$ calls to the weak learner to produce a classifier with (absolute) error at most $\varepsilon$ on any distribution consistent with the same labeling function. Thus, not only is Adaboost computationally efficient provided, of course, such weak learners can be found, but also its sample complexity is no worse than that of ERM. {\em This underscores the fact that the realizable boosting methodology has the potential to offer computational relief, without compromising on sample efficiency.}

\paragraph{Agnostic Boosting.} In practice, realizability is an onerous assumption; it is too limiting for the observed feature values alone to determine the label completely and deterministically, and that such a relation can be perfectly captured by an in-class hypothesis. Agnostic learning forgoes such assumptions. In their absence, bounds on absolute error are unachievable, e.g., when labels are uniformly random irrespective of features, no classifier can achieve accuracy better than half. Instead, in agnostic learning, the goal of the learner is to output a hypothesis with small {\em excess} error with respect to the best in-class classifier. If an in-class hypothesis is perfect on a given distribution, this relative error translates to an absolute error bound, thus generalizing the realizable case perfectly. Indeed, such model agnosticism has come to be a lasting hallmark of modern machine learning. 

Early attempts at realizing the promise of boosting in the agnostic setting were met with limited success: while they did boost the weak learner's accuracy, the final hypothesis produced was not competitive with the best in-class hypothesis. We survey some of these in the related work section. A later result, and the work most related to ours, is due to \cite{kanade2009potential}. A weak learner in this setting returns a classifier with a correlation against the true labels that is $\gamma$ (say $0.1$) times that of the best in-class hypothesis. Random guesses of the labels produce a correlation of zero; hence, a weak classifier interpolates the performance of the best in-class hypothesis with that of a random one. Given access to such a weak learner, the boosting algorithm of \cite{kanade2009potential} makes $\approx 1/\varepsilon^2$ calls to the weak learner and draws $m_{\text{agnostic boosting}}\approx (\log |\mathcal{H}|)/\varepsilon^4$ samples to produce a learning rule (not necessarily in the hypotheses class, hence {\em improper}) with $\varepsilon$-excess error. The dependency of the sample complexity in the target accuracy is thus quadratically worse than that of ERM. This gap persists untarnished for other known agnostic boosting algorithms too. {\em In this work, we seek to diminish this fundamental gap and construct a more sample-efficient agnostic boosting algorithm.}

Our main result is an efficient boosting algorithm that upon receiving $\approx (\log |\mathcal{H}|)/\varepsilon^3$ samples produces an {\em improper} learning rule with $\varepsilon$-excess error on the population loss. This is accomplished by the careful reuse of samples across rounds of boosting. We also extend these guarantees to infinite function classes, and give applications of our main result in reinforcement learning, and in agnostically learning halfspaces, in each case improving known results. We detail key contributions and technical challenges in achieving them next.

\subsection{Contributions and technical innovations}\label{sec:contrib}
\begin{table*}
    \begin{tabular}{@{}p{\textwidth}@{}}
        \centering
        \bgroup
        \def\arraystretch{2}
        \begin{tabular}{|c|c|c|}
            \hline
            &
            Episodic model 
            &
            Rollouts w. $\nu$-resets
            
            \\
            \hline
            \cite{brukhim2022boosting}
            & $1/\gamma^4\varepsilon^5$ & $1/\gamma^4\varepsilon^6$ 
            \\
            \hline
             \Cref{thm:rl} & $1/\gamma^3\varepsilon^4$               & $1/\gamma^3\varepsilon^5$  
            \\
            \hline
        \end{tabular}
        \egroup
    \end{tabular}
    \caption{Sample complexity of reinforcement learning given $\gamma$-weak learner over the policy class, for two different modes of accessing the underlying MDP, in terms of  $\varepsilon$ and $\gamma$,
    suppressing other terms.}
    \label{tab:table2}
\end{table*}

{\bf Contribution 1: Sample-efficient Agnostic Booster.} We provide a new potential-based agnostic boosting algorithm (\Cref{alg:rev}) that achieves $\varepsilon$-excess error when given a $\gamma$-weak learner operating with a base class $\mathcal{B}$. In \Cref{thm:maincons}, we prove that the sample complexity of this new algorithm scales as $(\log |\mathcal{B}|)/\gamma^3\varepsilon^3$, improving upon all known approaches to agnostic boosting. See \Cref{tab:table1}.

A key innovation in our algorithm design and the source of our sample efficiency is the careful recursive reuse of samples between rounds of boosting, via a second-order estimator of the potential in Line 5.II. In contrast, \cite{kanade2009potential} draws fresh samples for every round of boosting.

A second coupled innovation, this time in our analysis, is to circumvent a uniform convergence argument on {\em the boosted hypothesis class}, which would otherwise result in a sample complexity that scales as $\approx 1/\varepsilon^4$. Indeed, the algorithm in \cite{brukhim2020online} reuses the entire training dataset across all rounds of boosting. This approach succeeds in boosting the accuracy on the {\em empirical} distribution; however, success on the {\em population} distribution now relies on a uniform convergence (or sample compression) argument on the boosted class, the complexity of which grows linearly with the number of rounds of boosting since boosting algorithms are inevitably improper (i.e., output a final hypothesis by aggregating weak learners, hence outside the base class). Instead, we use a martingale argument on {\em the smaller base hypothesis class} to show that the empirical distributions constructed by our data reuse scheme and fed to the weak learner track the performance of any base hypothesis on the population distribution. This is encapsulated in \Cref{lem:vr}.



Finally, while we follow the potential-based framework laid in \cite{kanade2009potential}, we find it necessary to alter the branching logic dictating what gets added to the ensemble at every step. At each step, the algorithm makes progress via including the weak hypothesis or making a step ``backward'' via adding in a negation of the sign of the current ensemble. We note that there is a subtle error in \cite{kanade2009potential} (see \Cref{app:error}), that although for their purposes is rectifiable without a change in the claimed sample complexity, leads to $1/\varepsilon^4$ sample complexity here in spite of the above modifications. At the leisure of $1/\varepsilon^4$ sample complexity, the fix is to test which of these alternatives fares better by drawing fresh samples every round. However, given a smaller budget, the error of the negation of the sign of the ensemble, which lies outside the base class, is not efficiently estimable. Instead, in Line 9 we give a different branching criteria that can be evaluated using the performance of the weak hypothesis on past data alone.

{\bf Contribution 2: Trading off Sample and Oracle Complexity.}  Although \Cref{thm:maincons} offers an unconditional improvement on the sample complexity, it makes more calls to the weak learners than previous works. To rectify this, we give a second guarantee on the performance of \Cref{alg:rev} in \Cref{thm:maincons2} (in \Cref{app:improv}), with the oracle complexity matching that of known results. The resultant sample complexity improves known results for all practical (i.e., sub-constant) regimes of $\varepsilon$. This is made possible by a less well known variant of Freedman's inequality \citep{pinelis1994optimum, 371436} that applies to random variables bounded with high probability. 

{\bf Contribution 3: Extension to Infinite Classes.} Although in our algorithm the samples fed to the weak leaner are independent conditioned on past sources of randomness, our relabeling and data reuse across rounds introduces complicated inter-dependencies between samples. For example, a sample drawn in the past can simultaneously be used as is in the present round (i.e., in Line 5.I), in addition to implicitly being used to modify the label of a different sample via the weak hypothesis it induced in the past (i.e., in Line 5.II). Thus, the textbook machinery of extending finite hypothesis results to infinite class applicable to IID samples via symmetrization and Rademacher complexity (see, e.g., \cite{shalev2014understanding,mohri2018foundations,boucheron2005theory}) is unavailable to us. Instead, we first derive $L_1$ covering number based bounds in \Cref{thm:maincover} (in \Cref{app:inf}). Through a result in empirical process theory \citep{van1997weak}, we translate these to a $\text{VC}(\mathcal{B})/\gamma^3\varepsilon^3$ sample complexity bound, where $\text{VC}(\mathcal{B})$ is the VC dimension of class $\mathcal{B}$, in \Cref{thm:mainvc}. 

{\bf Contribution 4: Applications in Reinforcement Learning and Agnostic Learning of Halfspaces.} Building on earlier reductions from reinforcement learning to supervised learning \citep{kakade2002approximately}, \cite{brukhim2022boosting} initiated the study of function-approximation compatible reinforcement learning, given access to a weak learner over the policy class. By applying our agnostic booster to this setting, we improve their sample complexity by $\text{poly}(\varepsilon,\gamma)$ factors for binary-action MDPs, as detailed in \Cref{tab:table2}. Also, following \cite{kanade2009potential}, we apply our agnostic boosting algorithm to the problem of learning halfspaces over the Boolean hypercube and exhibit improved boosting-based results in \Cref{thm:half}.

{\bf Contribution 5: Experiments.} In preliminary experiments in \Cref{sec:exp}, we demonstrate that the sample complexity improvements of our approach do indeed manifest in the form of improved empirical performance over previous agnostic boosting approaches on commonly used datasets. 

\section{Related work}
The possibility of boosting was first posed in \cite{kearns1994cryptographic}, and was resolved positively in a remarkable result due to \cite{schapire1990strength} for the realizable case. The Adaboost algorithm \cite{freund1997decision} paved the way for its practical applications (notably in \cite{viola2001rapid}). We refer the reader to \cite{schapire2013boosting} for a comprehensive text that surveys  the many facets of boosting, including its connections to game theory and online learning. See also \cite{hogsgaard23a, green2022optimal, alon2021boosting} for recent developments. 

The fact that Adaboost and its natural variants are brittle in presence of label noise and lack of realizability \citep{long2008random} prompted the search for boosting algorithms in the realizable plus label noise \citep{diakonikolas2021boosting, kalai2003boosting} and agnostic learning models \citep{ben2001agnostic, mansour2002boosting, kalai2008agnostic, kale2007boosting, chen2016communication}. In general, these boosting models are incomparable: although agnostic learning implies success in the random noise model, agnostic weak learning also constitutes a stronger assumption. Early agnostic boosting results could not boost the learner's accuracy to match that of the best in-class hypothesis; this limitation was tied to their notion of agnostic weak learning. Our work is most closely related to \cite{kanade2009potential, feldman2009distribution}; we use the same notion of agnostic weak learning. 

Boosting has also been extended to the online setting. \cite{beygelzimer2015optimal, chen2012online, jung2017online} study boosting in the mistake bound (realizable) model, while \cite{brukhim2020online, raman2022online, hazan2021boosting} focus on regret minimization. Our scheme of data reuse is inspired by variance reduction techniques \cite{schmidt2017minimizing,johnson2013accelerating,fang2018spider,pmlr-v201-agarwal23a} in convex optimization, although there considerations of uniform convergence and generalization are absent, and our algorithm does not admit a natural gradient descent interpretation.


\section{Problem setting}
Let $\mathcal{D}\in \Delta(\mathcal{X}\times \{\pm 1\})$ be the joint population distribution over features, chosen from $\mathcal{X}$, and (signed) binary labels with respect to which a classifier's $h:\mathcal{X}\to \{\pm 1\}$ performance may be assessed. The performance criterion we consider is the 0-1 loss over the true labels and the classifier's predictions. 
$$ l_\mathcal{D}(h) = {\mathbb{E}}_{(x,y)\sim \mathcal{D}} \left[\mathds{1}(h(x)\neq y)\right]={\rm Pr}_{(x,y)\sim \mathcal{D}} [h(x)\neq y] $$
Relatedly, one may measure the correlation between the classifier's predictions and the true labels. 
$$ \corr_\mathcal{D}(h) = {\mathbb{E}}_{(x,y)\sim \mathcal{D}} \left[yh(x) \right] $$
Note that for signed binary labels, we have that $l_\mathcal{D}(h) = \frac{1}{2}(1-\corr_\mathcal{D}(h))$. Therefore, these notions are equivalent in that a classifier that maximizes the correlation with true labels also minimizes the 0-1 loss, and vice versa, even in a relative error sense.
\begin{definition}[Agnostic Weak Learner]\label{def:wl}
A learning algorithm is called a $\gamma$-agnostic weak learner with sample complexity $m:\mathbb{R}_+\times \mathbb{R}_+ \to \mathbb{N}\cup \{+\infty\}$ with respect to a hypothesis class $\mathcal{H}$ and a base hypothesis class $\mathcal{B}$ if, for any $\varepsilon_0, \delta_0>0$, upon being given $m(\varepsilon_0, \delta_0)$ independently and identically distributed samples from any distribution $\mathcal{D}'\in \Delta(\mathcal{X}\times \{\pm 1\})$, it can output a base\footnote{Typically, the base class $\mathcal{B}$ is (often substantially) smaller than $\mathcal{H}$. For example, decision stumps are a common example of the base class.} hypothesis $\mathcal{W}\in \mathcal{B}$ such that with probability $1-\delta_0$ we have
$$ \corr_\mathcal{D'}(\mathcal{W}) \geq \gamma \max_{h\in \mathcal{H}}\corr_\mathcal{D'}(h) -\varepsilon_0.$$
\end{definition} 

As remarked in \cite{kanade2009potential}, typically $m(\varepsilon,\delta) = O(({\log |\mathcal{B}|/\delta})/{\varepsilon^2})$, and we use this fact in compiling \Cref{tab:table1}. However, following \cite{kanade2009potential, brukhim2020online}, we state our main result for fixed $\varepsilon_0, \delta_0$ in \Cref{thm:maincons}, where a necessary and irreducible $\varepsilon_0$ term shows up in our final accuracy.

Although not explicitly mentioned in our weak learning definition, our algorithm falls within the {\em distribution-specific} boosting framework \citep{kanade2009potential, feldman2009distribution}. In particular, like previous work on agnostic boosting, \Cref{alg:rev} can be implemented by {\em relabeling examples}, instead of adaptively reweighing them. Thus, the overall marginal distribution of any $\mathcal{D}'$ fed to the learner on the feature space $\mathcal{X}$ is the same as that induced by the population distribution $\mathcal{D}$ on $\mathcal{X}$. Under such promise on inputs, distribution-specific weak learners may be easier to find.

\section{The algorithm and main results}

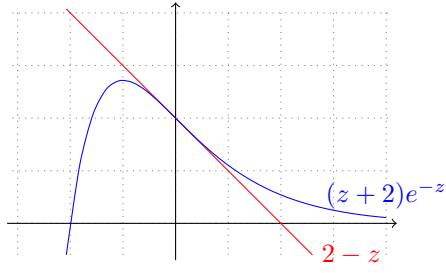
\begin{figure}
	\centering
    \begin{tikzpicture}[scale=0.7]
        \draw[color=gray, thin, dotted] (-3.1, -0.6) grid (4.1, 4.1);
    
        \draw[->] (-3.2,0) -- (4.2,0);
        \draw[->] (0,-0.7) -- (0,4.2);
    
        \draw[color=red, domain=-2.075:2.6] plot (\x, {2 - \x}) node[right] {$2 - z$};
        \draw[color=blue, domain=-2.075:4, smooth] plot (\x, {(\x + 2) * exp(-\x)}) node[above] {$(z + 2) e^{-z}$};
    \end{tikzpicture}
	\caption{The two components of the piecewise potential function $\phi(z)$, with $\color{blue}(z+2)e^{-z}$ plotted in {\color{blue} blue}, and $\color{red} 2-z$ in {\color{red} red}. Note that $\phi(z)$ is the point-wise maximum of the two.}
\end{figure}

{\bf Notations.} Given two functions $f,g:\mathcal{X}\to \mathbb{R}$ and generic scalars $\alpha,\beta\in \mathbb{R}$, we use $af+bg$ to denote a function such that $(\alpha f+\beta g)(x) = \alpha f(x)+\beta g(x)$ for all $x\in \mathcal{X}$. Given a function $f:\mathcal{X}\to \mathbb{R}$, we take $\sign(f)$ to be a function such that for all $x\in \mathcal{X}$, $\sign(f)(x) = \mathds{1}({f(x)\geq 0})-\mathds{1}({f(x)< 0})$. Define a filtration sequence $\{\mathcal{F}_t: t\in \mathbb{N}_{\geq 0}\}$, where $\mathcal{F}_t$ capture all source of randomness the algorithm is subject to in the first $t$ iterations. For brevity, we define $\mathbb{E}_{t}[\cdot] = \mathbb{E}[\cdot |\mathcal{F}_t]$. For any feature-label dataset $\widehat{D}$, we use $\mathbb{E}_{\widehat{D}}$ and $\corr_{\widehat{D}}$ to denote the empirical average and empirical correlation over $\widehat{D}$.

\begin{algorithm}[ht]
\begin{algorithmic}[1]
    \STATE \textbf{Inputs:} Sampling oracle for $\mathcal{D}$ supported on $\mathcal{X}\times \{\pm 1\}$, $\gamma$-agnostic weak learning oracle $\mathcal{W}$, step-size $\eta$, mixing parameter $\sigma$, number of iterations $T$, per-iteration sample size $S$, resampling parameter $m$, branching tolerance $\tau$, post-selection sample size $S_0$, potential $\phi:\mathbb{R}\to\mathbb{R}$. 
    \STATE Initialize a zero hypothesis $H_1=\mathbf{0}$.
    \FOR{$t=1$ to $T$}
        \STATE Sample $S$ IID examples from the distribution $\mathcal{D}$ to create a dataset $\widehat{D}_t$.
        \STATE Construct a sampling distribution $\mathcal{D}_t$ that samples $(x,y)$ uniformly from $\widehat{D}_t$ if $t=1$,\\ and for $t>1$ produces IID samples $(x,\hat{y})$ as follows:
            \begin{enumerate}[label=\Roman*]
            	\item With probability $1-\sigma$, return a sample $(x,\hat{y})$ from $\mathcal{D}_{t-1}$.
            	\item With remaining probability $\sigma$, draw $\eta' \sim \text{Unif}[0,\eta]$, pick $(x,y)$ uniformly from $\widehat{D}_t$, \\construct a pseudo label
            	$\hat{y} =\begin{cases}
            		+1 & \text{with probability } p_t(x,y,\eta'),\\
            		-1 & \text{with probability } 1-p_t(x,y,\eta'),
            	\end{cases}$ \\
            	and return $(x, \hat{y})$, where
            	$$p_t(x,y,\eta')= \frac{1}{2}-\frac{\sigma\phi'(yH_{t-1}(x))y + \eta\phi''(y(H_{t-1}(x)+\eta' h_{t-1}(x))) h_{t-1}(x)}{2(\eta+\sigma)}.$$
            \end{enumerate}
        \STATE Sample $m$ samples from $\mathcal{D}_t$ to create another dataset $\widehat{D}'_t$.
        \STATE Call the weak learning oracle on $\widehat{D}'_t$ to get $W_t=\mathcal{W}(\widehat{D}'_t)$.
        \STATE Measure the empirical correlation of $W_t$ on $\widehat{D}'_t$  as $\corr_{\widehat{D}'_t}(W_t) = \sum_{(x,\widehat{y})  \in \widehat{D}'_t} \widehat{y} W_t(x) $.
        \STATE Set $h_t= W_t/\gamma$ if $\corr_{\widehat{D}'_t}(W_t)>\tau$ else $h_t =-\sign(H_t)$.
        \STATE Update $H_{t+1} = H_t + \eta h_t$.
    \ENDFOR
    \STATE Sample $S_0$ IID examples from the distribution $\mathcal{D}$ to create a dataset $\widehat{D}_0$.
    \STATE \textbf{Output} the hypothesis $\overline{h} = \arg\max_{h\in \{\sign(H_t):t\in [T]\}} \sum_{(x,y)\in \widehat{D}_0} yh(x)$.
    \caption{Agnostic Boosting via Sample Reuse} \label{alg:rev}
\end{algorithmic}
\end{algorithm}

{\bf Potential function.} Define the potential function $\phi:\mathbb{R}\to\mathbb{R}$ as 
\begin{equation}
	 \phi(z) = \begin{cases}
	2-z & \text{if } z\leq 0, \\
	(z+2)e^{-z} & \text{if } z>0.
\end{cases}\label{eq:phi}
\end{equation} 

We can use this to assign a population potential to any real-valued hypothesis $H:\mathcal{X}\to \mathbb{R}$ as 
$$ \Phi_\mathcal{D}(H) = {\mathbb{E}}_{(x,y)\sim \mathcal{D}} \left[\phi(yH(x)\right]. $$
To maximize correlation between true labels and the hypothesis $H$'s outputs, one wants $H(x)>0$ whenever $y=+1$ for most samples drawn from the underlying distribution. Since $\phi$ is a monotonically decreasing function, equivalently, higher classifier accuracy typically corresponds to lower values of population potential. However, there are limits to the utility of this argument: a low value of the potential alone does not translate to successful learning. In agnostic learning, one is concerned not with the error of the learned classifier {\em per se}, but with its excess error over the best in-class hypothesis. We will provide a precise relation between the potential and the excess error in \Cref{lem:cons}. 

We will use the following properties of $\phi$ (proved in \Cref{app:aux}). Going forward, these will be the sole characteristics of $\phi$ we will appeal to. The potential we use is similar to the one used in \cite{kanade2009potential,domingo2000madaboost}, but has been modified to remove a jump discontinuity in the second derivative at $z=0$ as our approach requires a twice continuously differentiable potential. 
\begin{lemma}\label{lem:phi}
We make the following elementary observations about $\phi$:
\begin{enumerate}[nosep,label=\Roman*.]
	\item $\phi$ is convex and in $\mathcal{C}^2$, i.e., it is two-times continuously differentiable everywhere.
	\item $\phi$ is non-negative on $\mathbb{R}$ and $\phi(0) = 2$.
	\item For all $z\in \mathbb{R}$, $\phi'(z) \in [-1, 0]$. Further, for any $z<0$, $\phi'(z)=-1$.
	\item $\phi$ is $1$-smooth, i.e., $\forall z\in \mathbb{R}$, $\phi''(z) \leq 1$.
\end{enumerate}
\end{lemma}

For any real-valued hypotheses $H$, $h$ and $g$ on $\mathcal{X}$, to ease analysis, we introduce 
\begin{align*}
	\Phi'_\mathcal{D}(H, h) &= \mathbb{E}_{(x,y)\sim \mathcal{D}}[\phi'(yH(x))yh(x)],\\
	\Phi''_\mathcal{D}(H, h, g) &= \mathbb{E}_{(x,y)\sim \mathcal{D}}[\phi''(yH(x))h(x)g(x)].
\end{align*}
Equivalently, $\Phi'_\mathcal{D}(H, h)$ can be characterized as the $\mathcal{D}$-induced semi inner product between the functional derivative $\partial \Phi_\mathcal{D}(H)/\partial H$ and $h$. But for ease of presentation, we forgo this formal interpretation in favor of the literal one stated above.

A key property of the above potential is stated in the next lemma (proved in \Cref{app:aux}). It gives us a strategy to control the relative error of the learned hypothesis, the quantity on the right, by individually minimizing both terms on the left, as we discuss next.

\begin{lemma}\label{lem:cons}
For any distribution $\mathcal{D}\in \Delta(\mathcal{X}\times \{\pm 1\})$, real-valued hypothesis $H:\mathcal{X}\to\mathbb{R}$, and binary hypothesis $h^*:\mathcal{X}\to\{\pm 1\}$, we have
$$ \Phi'_\mathcal{D}(H, \sign(H)) - \Phi'_\mathcal{D}(H, h^*) \geq \corr_\mathcal{D}(h^*) - \corr_\mathcal{D}(\sign(H)).$$
\end{lemma}

{\bf Description of the algorithm.} Each round of \Cref{alg:rev} adds some multiple of either the weak hypothesis $W_t$ or $-\text{sign}(H_t)$ to the current ensemble $H_t$; this choice is dictated by the empirical correlation of the weak hypothesis on the dataset it was fed. The construction of the relabeled distribution $\mathcal{D}_t$ via our data reuse scheme ensures that if any hypothesis in the base class $\mathcal{B}$ produces sufficient correlation on it, its addition to the ensemble must decrease the potential $\Phi$ associated with the ensemble. Concretely, as we prove in \Cref{lem:vr}, for all $h\in \mathcal{B}$, $\text{corr}_{\mathcal{D}_t}(h)$ closely tracks $-\Phi'(H_t, h)$. The key to our improved sample complexity is the fact that this invariant can be maintained while sampling only $S\approx 1/\gamma\varepsilon$ fresh samples each round, by  repurposing samples from earlier rounds to construct the dataset fed to the weak learner. The mixing parameter $\sigma$ controls the proportion of these two sources of samples used to construct $\mathcal{D}_t$.

However, this explanation is opaque when it comes to motivating the need for $-\text{sign}(H_t)$. Let's rectify that: let $h^* \in \arg\min_{h\in \mathcal{H}}\text{corr}_\mathcal{D}(h)$ be the best in-class hypothesis. If $-\Phi'_\mathcal{D}(H_t, h^*)$ is sufficiently large, so is $\corr_{\mathcal{D}_t}(h^*)$ by \Cref{lem:vr}, which assures us a non-negligible weak learning edge.  As such $-\Phi'_\mathcal{D}(H_t, W_t)$ is large and the potential drops. If the weak hypothesis fails to make sufficient progress, the algorithms adds $-\sign(H_t)$ to the ensemble, which, again by \Cref{lem:vr} corresponds to decreasing the potential value by some function of $-\Phi'_\mathcal{D}(H_t, -\sign(H_t)) = \Phi'_\mathcal{D}(H_t, \sign(H_t))$, using linearity of $\Phi'_\mathcal{D}$ in its second argument. Thus, when run for sufficiently many iterations, because the potential is bounded above at initialization both the terms on the left side of \Cref{lem:cons} must become small. This then implies a bounded correlation gap between the best in-class hypothesis, and the majority vote of the ensembles considered in some iteration; the last line picks the best of these.

\subsection{Main result for finite hypotheses classes}
The key feature of our algorithm is that it is designed to reuse samples across successive rounds of boosting. The soundness of this scheme, which \Cref{lem:vr} substantiates, is based on the observation that in each round $H_t$ changes by a small amount, thereby inducing an incremental change in the distribution fed to the weak learner. Although our algorithm needs a total number of iterations comparable to \cite{kanade2009potential}, this data reuse lowers the number of fresh samples needed every iteration to just $1/\gamma\varepsilon$, instead of $1/\gamma^2\varepsilon^2$, resulting in improved sample complexity, as we show next. 

\begin{theorem}[Main result for finite hypotheses class]
\label{thm:maincons}
Choose any $\varepsilon,\delta>0$. There exists a choice of $\eta,\sigma,T, \tau, S_0, S, m$ satisfying\footnote{Here, the $\mathcal{O}$ notation suppresses polylogarithmic factors.} $T = \mathcal{O}((\log |\mathcal{B}|)/\gamma^2\varepsilon^2),  \eta = \mathcal{O}({\gamma^2 \varepsilon}/{\log |\mathcal{B}|}),  \sigma = {\eta}/{\gamma}, \tau = \mathcal{O}({\gamma\varepsilon}),   S= \mathcal{O}({1}/{\gamma\varepsilon}),  S_0 =\mathcal{O}({1}/{\varepsilon^2}), m=m(\varepsilon_0, \delta_0)+\mathcal{O}(1/\gamma^2\varepsilon^2)$
such that for any $\gamma$-agnostic weak learning oracle (as defined in \Cref{def:wl}) with fixed tolerance $\varepsilon_0$ and failure probability $\delta_0$, Algorithm~\ref{alg:rev} when run with the potential defined in \eqref{eq:phi} 
produces a hypothesis $\overline{h}$ such that with probability $1-10\delta_0 T-10\delta T$,
	$$ \corr_\mathcal{D}(\overline{h}) \geq \max_{h\in \mathcal{H}}\corr_\mathcal{D}(h) - \frac{2\varepsilon_0}{\gamma} - \varepsilon, $$
while making $T= \mathcal{O}((\log |\mathcal{B}|)/\gamma^2\varepsilon^2)$ calls to the weak learning oracle, and sampling $TS+S_0=\mathcal{O}((\log |\mathcal{B}|)/\gamma^3\varepsilon^3)$ labeled examples from $\mathcal{D}$.
\end{theorem}


In \Cref{app:improv}, we provide a different result (\Cref{thm:maincons2}), also using \Cref{alg:rev}, where the learner makes $\mathcal{O}(1/\gamma^2\varepsilon^2)$ call to weak learner, exactly matching the oracle complexity of existing results, while drawing $\mathcal{O}((\log |\mathcal{B}|)/\gamma^3\varepsilon^3 + (\log |\mathcal{B}|)^3/\gamma^2\varepsilon^2)$ samples.

\subsection{Extensions to infinite classes}
As mentioned in \Cref{sec:contrib}, the reuse of samples prohibits us from appealing to symmetrization and Rademacher complexity based arguments. Instead, our generalization to infinite classes is based on $L_1$ covering numbers. Using empirical process theory \citep{van1997weak}, we upper bound $L_1$ covering number by a suitable function of VC dimension, yielding the following result (proved in \Cref{app:inf}).

\begin{theorem}[Main result for VC Dimension]\label{thm:mainvc}
There exists a setting of parameters such that for any for any $\gamma$-agnostic weak learning oracle with fixed tolerance $\varepsilon_0$ and failure probability $\delta_0$, Algorithm~\ref{alg:rev} 
produces a hypothesis $\overline{h}$ such that with probability $1-10\delta_0 T-10\delta T$,
	$$ \corr_\mathcal{D}(\overline{h}) \geq \max_{h\in \mathcal{H}}\corr_\mathcal{D}(h) - \frac{2\varepsilon_0}{\gamma} - \varepsilon, $$
while making $T= \mathcal{O}(\text{VC}(\mathcal{B})/\gamma^2\varepsilon^2)$ calls to the weak learning oracle, and sampling $TS+S_0=\mathcal{O}(\text{VC}(\mathcal{B})/\gamma^3\varepsilon^3)$ labeled examples from $\mathcal{D}$.
\end{theorem}

\section{Sketch of the analysis}
In this section, we provide a brief sketch of the analysis outlined in the proof of \Cref{thm:maincons}. Our intent is to convey the plausibility of a $1/\varepsilon^3$ sample complexity result, up to the exclusion of other factors. Hence, $\approx$ and $\lesssim$ inequalities below only hold up to constants and polynomial factors in other paramters, e.g., in $\gamma, \log |\mathcal{B}|$. Formal proofs are reserved for \Cref{app:main}. 

{\bf Bounding the correlation gap (\Cref{thm:maincons}).} A central  tool in bounding the correlation gap is \Cref{lem:cons}. We want to ensure for some $t$, since our algorithm at the end picks the best one, that
$$ \underbrace{-\Phi'_\mathcal{D}(H_t, -\sign(H_t))}_{\text{want } \lesssim \varepsilon} + \underbrace{(-\Phi'_\mathcal{D}(H_t, h^*))}_{\text{want }\lesssim \varepsilon} \geq \corr_\mathcal{D}(h^*) - \corr_\mathcal{D}(\sign(H_t)).$$
On the other hand, using $1$-smoothness of $\phi$, we can upper bound $\Phi_\mathcal{D}$ on successive iterates as
$
	\Phi_\mathcal{D}(H_{t+1}) \leq \Phi_\mathcal{D}(H_t) + \eta \Phi'_\mathcal{D}(H_t, h_t) + {\eta^2}/{2\gamma^2}
$. Rearranging this to telescope the sum produces
	\begin{align*}
		-\frac{1}{T}\sum_{t=1}^T\Phi'_\mathcal{D}(H_t, h_t) &\leq \frac{\sum_{t=1}^T(\Phi_\mathcal{D}(H_t) - \Phi_\mathcal{D}(H_{t+1}))}{\eta T}  + \frac{\eta}{2\gamma^2} \leq \frac{2}{\eta T} + \frac{\eta}{2\gamma^2} \nonumber
	\end{align*}
Hence, by setting a $\eta\approx 1/\sqrt{T}$, we know that there exists a $t$ where $-\Phi'_\mathcal{D}(H_t, h_t)\lesssim 1/\sqrt{T}$. 

In \Cref{lem:vr}, we establish that the core guarantee our data resue scheme provides: that for all $h$ in the base class $\mathcal{B}$ and for $h^*$, the correlation on the resampled distribution $\mathcal{D}_t$ constructed by the algorithm tracks the previously stated quantity of interest $-\Phi'_\mathcal{D}(H_t, \cdot)$. 
\begin{lemma}\label{lem:vr}
	There exists a $C>0$ such that with probability $1-\delta$, for all $t\in [T]$ and $h\in \mathcal{B}\cup \{h^*\}$, we have
	$$ \left\lvert \Phi'_\mathcal{D}(H_t, h) + \left(1+\frac{\eta}{\sigma}\right)\corr_{\mathcal{D}_t}(h)\right\rvert\leq \underbrace{C\left(\sigma + \frac{\eta}{\gamma}\right)\left(\frac{1}{\sqrt{\sigma S}}\sqrt{\log\frac{|\mathcal{B}|T}{\delta}} + \log \frac{|\mathcal{B}|T}{\delta}\right)}_{\coloneqq \varepsilon_{\text{Gen}}}.$$
\end{lemma}
Using the definition of weak learner, we know that $\text{corr}_{\mathcal{D}_t}(h^*) \lesssim {\text{corr}_{\mathcal{D}_t}(W_t)}/{\gamma} $.
Now, using \Cref{lem:vr} twice and the linearity of $\Phi'_\mathcal{D}(H,\cdot)$, we get 
$$ - \Phi'_\mathcal{D}(H_t, h^*) \lesssim \text{corr}_{\mathcal{D}_t}(h^*) + \varepsilon_{\text{Gen}} \lesssim {\text{corr}_{\mathcal{D}_t}(W_t)}/{\gamma} + \varepsilon_{\text{Gen}} \lesssim -\Phi'_\mathcal{D}(H_t, W_t/\gamma) + 2\varepsilon_{\text{Gen}}.$$
Now, if at each step we could choose $h_t \in \{-\sign(H_t), W_t/\gamma\}$, which ever maximized $-\Phi'_\mathcal{D}(H_t, \cdot)$, we would have for some $t$ that 
$$ \max\{ -\Phi'_\mathcal{D}(H_t, -\sign(H_t)), -\Phi'_\mathcal{D}(H_t, h^*)-2\varepsilon_{\text{Gen}}\} \leq  -\Phi'_\mathcal{D}(H_t, h_t)\lesssim 1/\sqrt{T}. $$ 

Alas, $-\sign(H_t)$ is not in $\mathcal{B}$, thus $\text{corr}_{\mathcal{D}_t}(-\sign(H_t))$ can be really far from $-\Phi'_\mathcal{D}(H_t, -\sign(H_t))$, i.e., \Cref{lem:vr} does not apply to $-\text{sign}(H_t)$. To circumvent this, instead of choosing the maximizer, in the algorithm and in the actual proof, we use a relaxed criteria for choosing between $W_t/\gamma$ and $-\text{sign}(H_t)$ that depends on the correlation of $W_t$ on $\mathcal{D}_t$ alone, and hence can be efficiently evaluated. The spirit of this modification is to adopt $-\text{sign}(H_t)$ only if $W_t$ by itself fails to make enough progress, the threshold for which can be stated in the terms of target accuracy.


{\bf Generalization over $\mathcal{B}$ via sample reuse (\Cref{lem:vr}).} Here, we sketch a proof of \Cref{lem:vr} that ties the previous proof sketch together, and forms the basis of our sample reuse scheme. Our starting point is the following claim (\Cref{cl:phiexp}) that uses the fact that $\phi$ is second-order differentiably continuous to arrive at the fact that for any $t$ and $h:\mathcal{X}\to \mathbb{R}$, we have 
$$ \Phi'_\mathcal{D}(H_t, h) \approx \Phi'_\mathcal{D}(H_{t-1}, h) + \eta \Phi''_\mathcal{D}(H_{t-1}, h, h_{t-1}). $$
Simultaneously, using $\mathbb{E}_{t-1}$ to condition on the randomness in the first $t-1$ rounds, by construction, our data reuse and relabeling scheme gives us 
\begin{align*}
\mathbb{E}_{t-1}[\text{corr}_{\mathcal{D}_t}(h)]  \approx (1-\sigma)\text{corr}_{\mathcal{D}_{t-1}}(h)  - \frac{\sigma^2}{\sigma+\eta} \Phi'_\mathcal{D}(H_{t-1}, h) - \frac{\eta\sigma}{\sigma+\eta} \Phi''_\mathcal{D}(H_{t-1}, h, h_{t-1}).
\end{align*}
Thus, suitably scaling and adding the two together, we get the identity that 
	\begin{align*}
		\mathbb{E}_{t-1}[\Delta_t] &= \Phi'_\mathcal{D}(H_t, h) + \left(1+\frac{\eta}{\sigma} \right) \mathbb{E}_{t-1} \left[\text{corr}_{\mathcal{D}_t}(h)\right] \\
		&= (1-\sigma) \Phi'_\mathcal{D}(H_{t-1}, h) + (1-\sigma)\left(1+\frac{\eta}{\sigma}\right) \text{corr}_{\mathcal{D}_{t-1}}(h) = (1-\sigma) \Delta_{t-1},
	\end{align*}
where $\Delta_t = \Phi'_\mathcal{D}(H_t, h) + \left(1+\frac{\eta}{\sigma}\right)  \text{corr}_{\mathcal{D}_t}(h)$. Thus, $\Delta_t$ forms a martingale-like sequence; the sign of $\Delta_t$ is indeterminate. To establish concentration, we apply Freedman's inequality, noting that the conditional variance of the associated martingale difference sequence scales as $1/\sqrt{S}$. A union bound over $\mathcal{B}\cup \{h^*\}$ yields the claim. Relatedly, to reach a better oracle complexity in \Cref{thm:maincons2}, we show a uniform high-probability on the martingale difference sequence, and then apply a variant of Freedman's inequality \citep{pinelis1994optimum,371436} that can adapt to martingale difference sequences that are bounded with high probability instead of almost surely.

\section{Applications}
In this section, we detail the implications of our results for previously studied learning problems.

\subsection{Boosting for reinforcement learning}\label{sec:rl}
\cite{brukhim2022boosting} initiated the approach of boosting weak learners to construct a near-optimal policy for reinforcement learning. Plugging \Cref{alg:rev} into their meta-algorithm yields the following result for binary-action MDPs, improving upon the sample complexity in \cite{brukhim2022boosting}. Here, $V^\pi$ is the expected discounted reward of a policy $\pi$, $V^*$ is its maximum. $\beta$ is the discount factor of the underlying MDP, and $C_\infty,D_\infty$ and $\mathcal{E}, \mathcal{E}_\nu$ are distribution mismatch and policy completeness terms (related to the inherent Bellman error). In the {\em episodic model}, the learner interacts with the MDP in episodes. In the {\em $\nu$-reset model}, the learner can seed the initial state with a fixed well dispersed distribution $\nu$ as a means to exploration. See \Cref{app:rl} for a complete statement of results and details of the setting. 

\begin{theorem}[Informal; stated formally in \Cref{thm:rl2}]\label{thm:rl}
Let $\mathcal{W}$ be a $\gamma$-weak learner for the policy class $\Pi$ operating with a base class $\mathcal{B}$, with sample complexity $m(\varepsilon_0, \delta_0) = (\log |\mathcal{B}|/\delta_0)/\varepsilon_0^2$. Fix tolerance $\varepsilon$ and failure probability $\delta$. In the {\em episodic} access model, there is an algorithm using that uses the weak learner $\mathcal{W}$ to produce a policy $\overline{\pi}$ such that with probability $1-\delta$, we have
$ V^* - V^{\overline{\pi}} \leq {(C_\infty \mathcal{E})}/{(1-\beta)} + \varepsilon,$
while sampling $\mathcal{O}((\log |\mathcal{B}|)/\gamma^3 \varepsilon^4)$ episodes of length $\mathcal{O}((1-\beta)^{-1})$. In the {\em $\nu$-reset} access model, there is a setting of parameters such that \Cref{alg:rlMAIN1} when given access to $\mathcal{W}$ produces a policy $\overline{\pi}$ such that with probability $1-\delta$, we have
$ V^* - V^{\overline{\pi}} \leq {(D_\infty \mathcal{E}_\nu)}/{(1-\beta)^2} + \varepsilon$, while sampling $\mathcal{O}((\log |\mathcal{B}|)/\gamma^3 \varepsilon^5)$ episodes of length $\mathcal{O}((1-\beta)^{-1})$. 
\end{theorem}

\subsection{Agnostically learning halfspaces}\label{sec:halfspace}
We apply our algorithm in a black-box manner to agnostically learn halfspaces over the $n$-dimensional boolean hypercube when the data distribution has uniform marginals on features. The aim is this section is not to obtain the best known bounds, but rather to provide an example illustrating that agnostic boosting is both a viable and flexible approach to construct agnostic learners, and where our improvements carry over. Following \cite{kanade2009potential}, we use ERM over the parities of degree at most $d$, for  $d\approx 1/\varepsilon^4$, as our weak learners; the the weak learner's edge here is $\gamma = n^{-d}$. An application of our boosting algorithm (proved in \Cref{app:half}) to this problem improves the sample complexity of $\mathcal{O}(\varepsilon^{-8} n^{80\varepsilon^{-4}})$ indicated in \cite{kanade2009potential}.

\begin{theorem}\label{thm:half}
 Let $\mathcal{D}$ be any distribution over $\{\pm 1\}^n \times \{\pm 1\}$ with uniform distribution over features. By $\mathcal{H} = \{\sign(w^\top x - \theta) : (w,\theta) \in \mathbb{R}^{n+1}\}$, denote the class of halfspaces. There exists some $d$ such that running Algorithm~\ref{alg:rev} with ERM over parities of degree at most $d$ produces a classifier $\overline{h}$ such that $l_{\mathcal{D}}(\bar{h})  \leq  \min_{h \in \mathcal{H}} l_{\mathcal{D}}(h) + \varepsilon$, 
while using $\mathcal{O}(\varepsilon^{-7} n^{60\varepsilon^{-4}})$ samples in $n^{\text{poly}(1/\varepsilon)}$ time.
\end{theorem}

Note that ERM over the class of halfspaces directly, although considerably more sample efficient, takes $(1/\varepsilon)^{\text{poly}(n)}$ time, i.e., it is exponentially slower for moderate values of $\varepsilon$. There are known statistical query lower bounds \citep{diakonikolas2020near} requiring $n^{\text{poly}(\varepsilon^{-1})}$ queries for agnostic learning of halfspaces with Gaussian marginals, suggesting that a broad class of algorithms, regardless of the underlying parametrization, might not fare any better. For completeness, we note that a better sample complexity is attainable by direct $L_1$-approximations of halfspaces via low-degree polynomials \cite{diakonikolas2010bounded}, instead of the approach taken in \cite{kanade2009potential,klivans2004learning} and mirrored here which first constructs an $L_2$-approximation, but such structural improvements apply equally to presented and compared results.

\section{Experiments}\label{sec:exp}
In Table~\ref{tab:boosting_results}, we report the results of preliminary experiments with Algorithm~\ref{alg:rev} against the agnostic boosting algorithms in \cite{kanade2009potential} and \cite{brukhim2020online} as baselines on UCI classification datasets \citep{misc_ionosphere_52,misc_spambase_94,smith1988using}, using decision stumps \citep{scikit} as weak learners. We also introduce classification noise of 5\%, 10\% and 20\% during training to measure the robustness of the algorithms to label noise. Accuracy is estimated using $30$-fold cross validation with a grid search over the mixing weight $\sigma$ and the number of boosters $T$. The algorithm in \cite{kanade2009potential} does not reuse samples between rounds, while \cite{brukhim2020online} uses the same set of samples across all rounds. In contrast, \Cref{alg:rev} blends fresh and old samples every round, with $\sigma$ controlling the proportion of each. See Appendix~\ref{sec:exp_app} for additional details. We note that the Ionsphere dataset includes 351 samples, while Diabetes contains 768, and Spambase contains 4601. The benefits of sample reuse are less stark in a data-rich regime. This could explain some of the under-performance on Spambase, disregarding which the proposed algorithm substantially outperforms the alternatives.


\begin{table}[ht]
\centering
\scriptsize
\begin{tabular}{|c|c|c|c|c|c|c|}
\hline
Dataset & \multicolumn{3}{c|}{No Added Noise} & \multicolumn{3}{c|}{5\% Noise} \\
\hline
 & \cite{kanade2009potential} & \cite{brukhim2020online} & \textbf{Ours} &  \cite{kanade2009potential} & \cite{brukhim2020online} & \textbf{Ours} \\
\hline
Ionosphere & 0.92 $\pm$ 0.02 &  0.89 $\pm$ 0.03 & \textbf{0.97 $\pm$ 0.02} & 0.90 $\pm$ 0.03  & 0.88 $\pm$ 0.03& \textbf{0.97 $\pm$ 0.03} \\
\hline
Diabetes & 0.83 $\pm$ 0.03 &  0.78 $\pm$ 0.02 & \textbf{0.87 $\pm$ 0.03} & 0.83 $\pm$ 0.03  & 0.77 $\pm$ 0.02& \textbf{0.88 $\pm$ 0.03} \\
\hline
Spambase & 0.69 $\pm$ 0.02 & \textbf{0.79 $\pm$ 0.01} & 0.78 $\pm$ 0.02 & \textbf{0.81 $\pm$ 0.02} & 0.79 $\pm$ 0.01 & 0.78 $\pm$ 0.02  \\
\hline
German & 0.77 $\pm$ 0.02 & 0.75 $\pm$ 0.02 & \textbf{0.83 $\pm$ 0.02} & 0.78 $\pm$ 0.02 & 0.75 $\pm$ 0.02 & \textbf{0.85 $\pm$ 0.02} \\
\hline
Sonar & 0.66 $\pm$ 0.07 & \textbf{0.91 $\pm$ 0.03} &  0.88 $\pm$ 0.07 & 0.84 $\pm$ 0.05 & 0.88 $\pm$ 0.03 & \textbf{0.94 $\pm$ 0.05} \\
\hline
Waveform & 0.88 $\pm$ 0.01 &  0.78 $\pm$ 0.01 & \textbf{0.91 $\pm$ 0.01} & 0.88 $\pm$ 0.01  & 0.77 $\pm$ 0.01& \textbf{0.90 $\pm$ 0.01} \\
\hline
\hline
Dataset &  \multicolumn{3}{c|}{10\% Noise} & \multicolumn{3}{c|}{20\% Noise} \\
\hline
 & \cite{kanade2009potential} & \cite{brukhim2020online} & \textbf{Ours} &  \cite{kanade2009potential} & \cite{brukhim2020online} & \textbf{Ours}\\
\hline
Ionosphere &  0.93 $\pm$ 0.02  & 0.89 $\pm$ 0.02& \textbf{0.97 $\pm$ 0.02} & 0.92 $\pm$ 0.03  & 0.90 $\pm$ 0.03 & \textbf{0.96 $\pm$ 0.03}\\
\hline
Diabetes & 0.83 $\pm$ 0.03  & 0.78 $\pm$ 0.02& \textbf{0.88 $\pm$ 0.03} & 0.82 $\pm$ 0.02  & 0.78 $\pm$ 0.02 & \textbf{0.88 $\pm$ 0.02}\\
\hline
Spambase & \textbf{0.83 $\pm$ 0.02} & 0.80 $\pm$ 0.01 &  0.79 $\pm$ 0.02&  \textbf{0.84 $\pm$ 0.01}  & 0.79 $\pm$ 0.01& 0.79 $\pm$ 0.01 \\
\hline
German &  0.78 $\pm$ 0.02  & 0.75 $\pm$ 0.02 & \textbf{0.84 $\pm$ 0.02}& 0.78 $\pm$ 0.02  & 0.74 $\pm$ 0.02 & \textbf{0.84 $\pm$ 0.02}\\
\hline
Sonar & 0.85 $\pm$ 0.04 &  \textbf{0.91 $\pm$ 0.03} &0.88 $\pm$ 0.04 & 0.88 $\pm$ 0.04  & 0.88 $\pm$ 0.04& \textbf{0.93 $\pm$ 0.04} \\
\hline
Waveform & 0.88 $\pm$ 0.01 & 0.77 $\pm$ 0.01 & \textbf{0.91 $\pm$ 0.01} & 0.88 $\pm$ 0.01  & 0.77 $\pm$ 0.01& \textbf{0.90 $\pm$ 0.01} \\
\hline
\end{tabular}
\caption{Cross-validated accuracies of Algorithm \ref{alg:rev} compared to the agnostic boosting algorithms from \cite{kanade2009potential} and \cite{brukhim2020online} on 6 datasets. The first column reports accuracy on the original datasets, and the next three report performance with 5\%, 10\% and 20\% label noise added during training. The proposed algorithm simultaneously outperforms both the alternatives on 18 out of 24 instances.}
\label{tab:boosting_results}
\end{table}

\section{Conclusion}
We give an agnostic boosting algorithm with a substantially lower sample requirement than ones known, enabled by efficient recency-aware data reuse between boosting iterations. Improving our oracle complexity or proving its optimality, and closing the sample complexity gap to ERM are interesting directions for future work.

\nocite{*}
\bibliography{main}

\newpage
\appendix
\section*{Appendix}

\paragraph{Limitations.} The primary contribution of this work is theoretical. Extensively demonstrating the empirical efficacy of the proposed approach and fully characterizing when it fares best is left to future work. Further, in comparison to realizable boosting (with or without label noise), the existence of an agnostic weak leaner here is a stronger assumption. Nevertheless, we believe, especially given the recent interest in agnostic boosting, the current work presents a substantial and concrete improvement over the state of the art, and is a first step in making agnostic boosting practical.

\paragraph{Organization of the appendix.}
In \Cref{app:error}, we point out an error in the branching criteria in \cite{kanade2009potential}. Next, in \Cref{app:improv}, we give a second guarantee on our algorithm that improves the oracle complexity at a vanishing (in $\varepsilon$) cost to the sample complexity. In \Cref{app:aux,app:main}, we prove the main result and the lemmas leading up to it. \Cref{app:improvp} provides a proof of the improved result in \Cref{app:improv}. \Cref{app:inf} furnishes a proof for the claims concerning extensions to infinite classes. In \Cref{app:rl,app:rl2}, we define the reinforcement learning setup, formally state the RL boosting result along with accompanying algorithms, and prove it. \Cref{app:half} substantiates our improved sample complexity bound for learning halfspaces. Finally, in \Cref{sec:exp_app}, we provide additional experimental details, and \Cref{sec:guide_app} provides a practical guide for adapting the proposed algorithm.

\section{Branching criteria in \cite{kanade2009potential}}\label{app:error}
At a high level, the boosting algorithm in \cite{kanade2009potential} is an iterative one, that adds either the weak hypothesis or the negation of the sign of the present ensemble to the ensemble mixture in each of the $T\approx 1/\varepsilon^2$ rounds. The algorithm makes use of samples to fulfill two objectives: (a) provide $m\approx 1/\varepsilon^2$ samples to the weak learner to obtain a weak hypothesis that generalizes, (b) use $s\approx 1/\varepsilon^2$ samples to decide whether to add the weak hypothesis or the voting classifier to the current mixture, by comparing the empirical performance on both on said samples. The algorithm uses fresh samples for part (a) every round; this is sound and contributes $Tm \approx 1/\varepsilon^4$ to the net sample complexity.

However, as described in \cite{kanade2009potential}, the algorithm reuses the same $s$ samples for part (b) across all rounds. This means these $s$ samples determine which of the two choices gets added to the mixture at the end of the first step, and hence $H_1$. In the next step, however, because $H_1$ through relabelling of new samples determines the weak hypothesis, these samples are no longer IID with respect to the weak hypothesis or $-\sign(H_1)$, since they have already played a demonstrable part in determining it. This effects occur and compound at all time steps, not just the first. In the analysis, on top of page 7, the analysis in \cite{kanade2009potential} uses a Chernoff-Hoeffding bound to say that the performance of the weak hypothesis and $-\sign(H_1)$ on these $s$ samples transfers approximately to the population. However, this inequality may only be applied to IID random variables.

In the case of \cite{kanade2009potential}, there is a simple and satisfactory fix: resample these $s$ examples from the population distribution every round. The sample complexity, now $T(m+s)$ instead of $Tm+s$, remains $\mathcal{O}(1/\varepsilon^4)$, and no change to the performance gurantee needs to be made.

In our adaptation, this highlights an additional challenge. Even if one were to sate the weak learner with a total fewer number of samples, i.e., perform part (a) with fewer samples, through a uniform convergence result on the base hypothesis (crucially, not the {\em boosted hypothesis}, whose complexity grows with number of iterations), part (b) requires $s$ fresh samples, and hence $1/\varepsilon^4$ samples in total, in determining which of the weak hypothesis or $-\text{sign}$ of the ensemble provides a greater magnitude of descent. Here, note that $-\sign(H_t)$ lies outside the base class. To circumvent this, we give a branching criteria to decide which component to add to the present mixture, based on the performance of the weak hypothesis, the only component whose performance is estimable with bounded generalization error, alone on the reused data. Concretely, we introduce a threshold $\tau$ to choose the weak hypothesis, if it makes at least $\tau$ progress, and otherwise, choose $-\sign(H_t)$ even if in truth it is worse than the weak hypothesis. This deviation requires careful handling in the proof.

\section{Improved oracle complexity}\label{app:improv}
Here, we provide a different result where \Cref{alg:rev} makes $\mathcal{O}(1/\gamma^2\varepsilon^2)$ call to weak learner, matching exatcly the oracle complexity of existing results, while drawing $\mathcal{O}((\log |\mathcal{B}|)/\gamma^3\varepsilon^3 + (\log |\mathcal{B}|)^3/\gamma^2\varepsilon^2)$ samples. Notice that the second term in the sample complexity has a smaller order, whenever $\varepsilon$ is sub-constant. The proof may be found in \Cref{app:improvp}.

\begin{theorem}[Improved oracle complexity for finite hypotheses class]
\label{thm:maincons2}
Choose any $\varepsilon,\delta>0$. There exists a choice of $\eta,\sigma,T, \tau, S_0, S, m$ satisfying $T = \mathcal{O}(1/\gamma^2\varepsilon^2),  \eta = \mathcal{O}({\gamma^2 \varepsilon}),  \sigma = {\eta}/{\gamma}, \tau = \mathcal{O}({\gamma\varepsilon}),   S= \mathcal{O}({(\log |\mathcal{B}|)}/{\gamma\varepsilon}+(\log |\mathcal{B}|)^3),  S_0 =\mathcal{O}({1}/{\varepsilon^2}), m=m(\varepsilon_0, \delta_0)+\mathcal{O}(1/\gamma^2\varepsilon^2)$
such that for any $\gamma$-agnostic weak learning oracle (as defined in \Cref{def:wl}) with fixed tolerance $\varepsilon_0$ and failure probability $\delta_0$, Algorithm~\ref{alg:rev} when run with the potential defined in \eqref{eq:phi} 
produces a hypothesis $\overline{h}$ such that with probability $1-10\delta_0 T-10\delta T$,
	$$ \corr_\mathcal{D}(\overline{h}) \geq \max_{h\in \mathcal{H}}\corr_\mathcal{D}(h) - \frac{2\varepsilon_0}{\gamma} - \varepsilon, $$
while making $T= \mathcal{O}(1/\gamma^2\varepsilon^2)$ calls to the weak learning oracle, and sampling $TS+S_0=\mathcal{O}((\log |\mathcal{B}|)/(\gamma^3\varepsilon^3)+(\log |\mathcal{B}|)^3/(\gamma^2\varepsilon^2))$ labeled examples from $\mathcal{D}$.
\end{theorem}

\section{Proof of auxiliary lemmas}\label{app:aux}

\begin{proof}[Proof of \Cref{lem:phi}]
Part 2 is evident from the definition. Taking care that right and left first (and second) derivatives exist and are equal at $z=0$, by explicit computation, we have $$ \phi'(z) = \begin{cases}
	-1 & \text{if } z\leq 0, \\
	-(z+1)e^{-z} & \text{if } z>0,
\end{cases} \qquad \phi''(z) = \begin{cases}
	0 & \text{if } z\leq 0, \\
	ze^{-z} & \text{if } z>0.
\end{cases} $$
From this, one may immediately verify Part 1, since $\phi''$ is non-negative. Non-negativity of $\phi''$ also implies that $\phi'$ is non-decreasing, and hence, being clearly non-positive, is between $[-1,0]$; this is Part 3. By elementary calculus, $\phi''$ is maximize at $z=1$ where it equals $1/e$ implying the conclusion in Part 4. 
\end{proof}

\begin{proof}[Proof of \Cref{lem:cons}]
By the definition of $\Phi'_\mathcal{D}$, we have
\begin{align*}
	\Phi'_\mathcal{D}(H, \text{sign}(H)) - \Phi'_\mathcal{D}(H, h^*) =& \mathbb{E}_{(x,y)\sim \mathcal{D}} [\phi'(yH(x))y (\text{sign}(H)(x)-h^*(x))]\\
	=& \mathbb{E}_{(x,y)\sim \mathcal{D}} [\mathds{1}(yH(x)\geq 0)\phi'(yH(x))y (\text{sign}(H)(x)-h^*(x))] \\
	&+ \mathbb{E}_{(x,y)\sim \mathcal{D}} [\mathds{1}(yH(x)< 0)\phi'(yH(x))y (\text{sign}(H)(x)-h^*(x))].
\end{align*}
Note that whenever $yH(x)< 0$, $\phi'(yH(x)) = -1$, which \Cref{lem:phi}.III attests to. On the other hand, if $yH(x)\geq 0$, since $h^*$ is a binary classifier, $y\text{sign}(H)(x)=1\geq yh^*(x)$, and since, again by \Cref{lem:phi}.III, $\phi'(yH(x))\geq -1$, we have in this case 
$$ \phi'(yH(x)) y(\text{sign}(H)(x) - h^*(x)) \geq -y(\text{sign}(H)(x) - h^*(x))$$
Plugging these into the previous derivation, we arrive at
\begin{align*}
	\Phi'_\mathcal{D}(H, \text{sign}(H)) - \Phi'_\mathcal{D}(H, h^*) \geq & \mathbb{E}_{(x,y)\sim \mathcal{D}} [\mathds{1}(yH(x)\geq 0)y (h^*(x)-\text{sign}(H)(x))] \\
	&+ \mathbb{E}_{(x,y)\sim \mathcal{D}} [\mathds{1}(yH(x)< 0)y (h^*(x)-\text{sign}(H)(x))]\\
	=& \mathbb{E}_{(x,y)\sim \mathcal{D}} [y(h^*(x) - \text{sign}H(x))]\\
	=& \text{corr}_\mathcal{D}(h^*) - \text{corr}_\mathcal{D}(\text{sign}(H)), 
\end{align*}
finishing the proof of the claim.
\end{proof}

\section{Proofs for the main result}\label{app:main}

\begin{proof}[Proof of Theorem~\ref{thm:maincons}.]
Let $h^* \in \arg\min_{h\in \mathcal{H}}\text{corr}_\mathcal{D}(h)$. Using the update rule for $H_{t+1}$ and the fact that $\phi$ is $1$-smooth (\Cref{lem:phi}.IV), we arrive at 
	\begin{align*}
		\Phi_\mathcal{D}(H_{t+1})  &= \mathbb{E}_{(x,y)\sim \mathcal{D}} [\phi(y(H_{t}(x)+\eta h_t(x)))] \\
		&\leq \mathbb{E}_{(x,y)\sim \mathcal{D}} \left[\phi(yH_t(x)) + \eta y\phi'(yH_t(x))h_t(x) + (\eta h_t(x) y)^2/2\right]\\
		&\leq \Phi_\mathcal{D}(H_t) + \eta \Phi'_\mathcal{D}(H_t, h_t) + {\eta^2}/{2\gamma^2}, 
	\end{align*}
	using the definition of $\Phi'_{\mathcal{D}}$, and that $h_t$ is either a binary classifier, or a $1/\gamma$-scaled version of it. Rearranging this to telescope the sum produces
	\begin{align}\label{eq:smooth}
		-\frac{1}{T}\sum_{t=1}^T\Phi'_\mathcal{D}(H_t, h_t) &\leq \frac{\sum_{t=1}^T(\Phi_\mathcal{D}(H_t) - \Phi_\mathcal{D}(H_{t+1}))}{\eta T}  + \frac{\eta}{2\gamma^2} \nonumber \\
		&\leq \frac{2}{\eta T} + \frac{\eta}{2\gamma^2} 
	\end{align}
	where we use the fact that $\Phi_\mathcal{D}(\mathbf{0})=2$, and that it is non-negative (\Cref{lem:phi}.II).
 
 Hence, $\exists t\in [T]$, such that $-\Phi'_\mathcal{D}(H_t, h_t)$ is small. Our proof strategy going forward is to use this fact to imply that both the terms on left side of inequality in \Cref{lem:cons}, namely, $-\Phi_\mathcal{D}'(-\text{sign}(H_t))$ and $-\Phi'_\mathcal{D}(H_t, h^*)$, are small for some $t$, implying a small correlation gap on the population distribution. 

 Note that if $\text{corr}_{D'_t}(W_t) \geq \tau$, the algorithm sets $h_t = W_t/\gamma$, else it chooses $h_t = -\text{sign}(H_t)$. We analyze these cases separately. For both, \Cref{lem:vr} which relates the empirical correlation on $\mathcal{D}_t$ with $\Phi'_{\mathcal{D}}$ will prove indispensable. Going forward, we will define $\varepsilon_{\text{Gen}}$ as indicated above to capture the generalization error over the base hypothesis class.

For brevity of notation, we condition the analysis going forward on three events. Let $\mathcal{E}_A$ be that event that for all $t$, $|\text{corr}_{\mathcal{D}_t}(W_t) - \text{corr}_{{\widehat{D}}'_t}(W_t)|\leq\varepsilon'/10$. Since $\widehat{D}'_t$ is constructed from IID samples from $\mathcal{D}_t$, setting $m\geq {100}/{\varepsilon'^2}\sqrt{\log T/\delta}$, $\Pr(\mathcal{E}_A)\geq 1-\delta$, by an application of Hoeffding's inequality and union bound over $t$. Here, note that unlike $S$ setting a higher value of $m$ doesn't increase the number of points sampled from $\mathcal{D}$, since $D'_t$ is resampled from already collected data. Similarly, denote by $\mathcal{E}_B$ the event that for all $t$, $ |\text{corr}_\mathcal{D}(\text{sign}(H_t)) - \text{corr}_{\widehat{D}_0}(\text{sign}(H_t)) | \leq \varepsilon''/10$. Again, by Hoeffding's inequality and union bound over $t$, choosing $S_0= 100/\varepsilon''^2\sqrt{\log{T}/{\delta}}$, we have $\Pr(\mathcal{E}_B)\geq 1-\delta$, since the samples in $\widehat{D}_0$ were chosen independently of those used to compute $H_t$'s. Finally, we will take the success of \Cref{lem:vr} (call this $\mathcal{E}_C$) for granted in the analysis below, for brevity of notation, conditioning our analysis on all three events.

\paragraph{Case A: When $h_t=W_t/\gamma$.} When this happens, $\text{corr}_{\mathcal{D}_t}(W_t) \geq \text{corr}_{{D}'_t}(W_t)-\varepsilon'/10 \geq \tau -\varepsilon'/10$. Applying \Cref{lem:vr} and noting that $\Phi'_\mathcal{D}(H_t,\cdot)$ is linear in its argument, we have
\begin{align*}
	-\Phi'_\mathcal{D}\left(H_t, \frac{W_t}{\gamma}\right) &= -\frac{\Phi'_\mathcal{D}(H_t, W_t)}{\gamma} \\
	& \geq \frac{1}{\gamma}\left(1+\frac{\eta}{\sigma}\right) \text{corr}_{\mathcal{D}_t} (W_t) - \frac{\varepsilon_{\text{Gen}}}{\gamma} \\
	&\geq \frac{1}{\gamma}\left(1+\frac{\eta}{\sigma}\right)\left(\tau-\frac{\varepsilon'}{10}\right) - \frac{\varepsilon_{\text{Gen}}}{\gamma}
\end{align*}
Rearranging this 
\begin{align} \Phi'_\mathcal{D}\left(H_t, \frac{W_t}{\gamma}\right) \leq -\frac{1}{\gamma}\left(1+\frac{\eta}{\sigma}\right)\left(\tau-\frac{\varepsilon'}{10}\right) + \frac{\varepsilon_{\text{Gen}}}{\gamma}.\label{eq:casea}\end{align}

\paragraph{Case B: When $h_t=-\text{sign}(H_t)$.} Here $\text{corr}_{\mathcal{D}_t}(W_t) \leq \text{corr}_{{D}'_t}(W_t)+\varepsilon'/10 \leq \tau +\varepsilon'/10$. Applying \Cref{lem:vr}, and using the weak learning condition (\Cref{def:wl}), we have that
\begin{align*}
	\left(1+\frac{\eta}{\sigma}\right)\left(\tau + \frac{\varepsilon'}{10}\right) &\geq \left(1+\frac{\eta}{\sigma}\right)\text{corr}_{\mathcal{D}_t} (W_t)  \\
	&\geq \gamma \left(1+\frac{\eta}{\sigma}\right)\text{corr}_{\mathcal{D}_t} (h^*) -\left(1+\frac{\eta}{\sigma}\right)\varepsilon_0 \\
	&\geq -\gamma \Phi'_\mathcal{D}(H_t, h^*) -\left(1+\frac{\eta}{\sigma}\right)\varepsilon_0 - \gamma \varepsilon_{\text{Gen}}
\end{align*}
Now, we invoke the linearity of $\Phi'_\mathcal{D}(H_t, \cdot)$ and \Cref{lem:cons} to observe that 
\begin{align}
\Phi'_\mathcal{D}(H_t,-\text{sign}(H_t)) &= -\Phi'_\mathcal{D}(H_t,\text{sign}(H_t)) \nonumber\\
&\leq -	\Phi'_\mathcal{D}(H_t, h^*) -  (\text{corr}_\mathcal{D}(h^*)-\text{corr}_\mathcal{D}(\text{sign}(H_t)))\nonumber \\
&\leq  -(\text{corr}_\mathcal{D}(h^*)-\text{corr}_\mathcal{D}(\text{sign}(H_t))) + \frac{1}{\gamma}\left(1+\frac{\eta}{\sigma}\right)\left(\tau+\frac{\varepsilon'}{10}+\varepsilon_0\right) + \varepsilon_{\text{Gen}}.\label{eq:caseb}
\end{align}

\paragraph{Combining the two.} In either case, combining \Cref{eq:casea,eq:caseb}, we have 
\begin{align*}
\Phi'_\mathcal{D} ({H}_t, h_t ) \leq \max\bigg\{&  -\frac{1}{\gamma}\left(1+\frac{\eta}{\sigma}\right)\left(\tau-\frac{\varepsilon'}{10}\right) + \frac{\varepsilon_{\text{Gen}}}{\gamma} ,\\
&  -(\text{corr}_\mathcal{D}(h^*)-\text{corr}_\mathcal{D}(\text{sign}(H_t))) + \frac{1}{\gamma}\left(1+\frac{\eta}{\sigma}\right)\left(\tau+\frac{\varepsilon}{10}+\varepsilon_0\right) + \varepsilon_{\text{Gen}}\bigg\},
\end{align*}
or using the identity  $- \max (- f(x)) = \min f(x)$,
\begin{align*}
-\Phi'_\mathcal{D} ({H}_t, h_t ) \geq \min\bigg\{& \frac{1}{\gamma}\left(1+\frac{\eta}{\sigma}\right)\left(\tau-\frac{\varepsilon'}{10}\right) - \frac{\varepsilon_{\text{Gen}}}{\gamma} ,\\
&  (\text{corr}_\mathcal{D}(h^*)-\text{corr}_\mathcal{D}(\text{sign}(H_t))) - \frac{1}{\gamma}\left(1+\frac{\eta}{\sigma}\right) \left(\tau+\frac{\varepsilon'}{10}+\varepsilon_0\right) - \varepsilon_{\text{Gen}} \bigg\}.
\end{align*}
Now, set $$\tau = \left(1+\frac{\eta}{\sigma}\right)^{-1}\left(\frac{{4}}{\eta T} + \frac{\eta}{\gamma^2}+\frac{\varepsilon_{\text{Gen}}}{\gamma}\right) \gamma+ \frac{\varepsilon'}{10}.$$ 
Now, either there exists some $t$ such
\begin{align}\label{eq:calc1}
	 \text{corr}_\mathcal{D}(h^*)-\text{corr}_\mathcal{D}(\text{sign}(H_t)) &\leq \frac{1}{\gamma}\left(1+\frac{\eta}{\sigma}\right) (2\tau+\varepsilon_0) + \left(1-\frac{1}{\gamma}\right)\varepsilon_{\text{Gen}} \nonumber\\
	 &= \frac{8}{\eta T} + \frac{2\eta}{\gamma^2} + \left(1+\frac{1}{\gamma}\right)\varepsilon_\text{Gen} +\left(1+\frac{\eta}{\sigma}\right) \left(\frac{\varepsilon_0}{\gamma} + \frac{\varepsilon'}{5\gamma}\right),
\end{align}
or the minimum operator in the last expression always accepts the first clause, in which case for all $t$, $$-\Phi'_\mathcal{D} ({H}_t, h_t ) \geq \frac{1}{\gamma}\left(1+\frac{\eta}{\sigma}\right)\left(\tau-\frac{\varepsilon'}{10}\right) -\frac{\varepsilon_{\text{Gen}}}{\gamma} = \frac{{4}}{(\eta T)} + \frac{\eta}{\gamma^2},$$
which contradicts \Cref{eq:smooth}. Henceforth let ${t}^*$ be the iteration for which \Cref{eq:calc1} holds.

Finally, given the event $\mathcal{E}_B$, we have
\begin{align*}
\text{corr}_\mathcal{D} (\overline{h}) &\geq  \text{corr}_{\widehat{D}_0} (\overline{h}) -\frac{\varepsilon''}{10} \geq  \text{corr}_{\widehat{D}_0} ( \text{sign}(H_{{t}^*})) -\frac{\varepsilon''}{10} \geq \text{corr}_\mathcal{D}(\text{sign}(H_{{t}^*})) -\frac{\varepsilon''}{5}.
\end{align*}

Compiling this with the inequality in \Cref{eq:calc1}, we get
\begin{align*}
	 \text{corr}_\mathcal{D}(h^*)-\text{corr}_\mathcal{D}(\overline{h})\leq  \frac{8}{\eta T} + \frac{2\eta}{\gamma^2} + \frac{2\varepsilon_\text{Gen}}{\gamma} +\frac{1}{\gamma}\left(1+\frac{\eta}{\sigma} \right) \left(\varepsilon_0 + \frac{\varepsilon'}{5}\right) + \frac{\varepsilon''}{5}.
\end{align*}
Setting $\varepsilon' = \varepsilon/5\gamma$, $\varepsilon'' = \varepsilon/5$ and plugging in the proposed hyper-parameters with appropriate constants yields the claimed result.
\end{proof}

\begin{proof}[Proof of Lemma~\ref{lem:vr}.]
 Fix any hypothesis $h\in\mathcal{B}$. For any $\eta'\geq 0$, define $H_{t}^{\eta'} = H_t +\eta' h_t$, and 
	$$\Delta_t = \Phi'_\mathcal{D}(H_t, h) + \left(1+\frac{\eta}{\sigma}\right) \mathbb{E}_{(x,y)\sim \mathcal{D}_t} \left[yh(x)\right].$$   
First, we derive recursive expansions of $\Phi'_\mathcal{D}(H_t, h)$ and $\mathcal{D}_t$, the two quantities we wish to relate.
\begin{claim}\label{cl:phiexp}
For any $t$ and $h:\mathcal{X}\to \mathbb{R}$, we have 
$$ \Phi'_\mathcal{D}(H_t, h) = \Phi'_\mathcal{D}(H_{t-1}, h) + \eta \mathbb{E}_{\eta'\sim \text{Unif}[0,\eta]}[\Phi''_\mathcal{D}(H_{t-1}^{\eta'}, h, h_{t-1})]. $$
\end{claim}

\begin{claim}\label{cl:dexp1}
For any $t$,
\begin{align*}
&\mathbb{E}_{t-1}[\mathbb{E}_{(x,y)\sim \mathcal{D}_t} \left[yh(x)\right]] \\
& = (1-\sigma)\mathbb{E}_{(x,y)\sim \mathcal{D}_{t-1}} \left[yh(x)\right] - \frac{\sigma^2}{\sigma+\eta} \Phi'_\mathcal{D}(H_{t-1}, h) - \frac{\eta\sigma}{\sigma+\eta} \mathbb{E}_{\eta'\sim \text{Unif}[0,\eta]}[\Phi''_\mathcal{D}(H_{t-1}^{\eta'}, h, h_{t-1})].
\end{align*}
\end{claim}
	
Adding $(1+\eta/\sigma)$ times the last expression to the expansion of $\Phi'_\mathcal{D}(H_t, h)$, we have
	\begin{align*}
		\mathbb{E}_{t-1}[\Delta_t] &= \Phi'_\mathcal{D}(H_t, h) + \left(1+\frac{\eta}{\sigma} \right) \mathbb{E}_{t-1} \left[\mathbb{E}_{(x,y)\sim \mathcal{D}_t} \left[yh(x)\right] \right] \\
		&= (1-\sigma) \Phi'_\mathcal{D}(H_{t-1}, h) + (1-\sigma)\left(1+\frac{\eta}{\sigma}\right) \mathbb{E}_{(x,y)\sim \mathcal{D}_{t-1}} \left[yh(x)\right] \\
		&= (1-\sigma) \Delta_{t-1}.
	\end{align*}

	Using this, we conclude that $\Delta'_t = (1-\sigma)^{-t}\Delta_t$ forms a martingale sequence with respect to the $\{\mathcal{F}_t:t\in \mathbb{N}_{\geq 0}\}$ filtration sequence, as
	$ \Delta'_t = \mathbb{E}_{t-1}[(1-\sigma)^{-t} \Delta_t] = (1-\sigma)^{t-1} \Delta_{t-1} =\Delta'_{t-1} $. The associated martingale difference sequence $\delta'_t = \Delta'_t - \Delta'_{t-1}$ can be bounded both in worst-case and second-moment terms as we show next. 
	
\begin{claim}\label{cl:delta1}
For all $t$, $|\delta'_t|\leq (1-\sigma)^{-t} 2(\sigma + \eta/\gamma)$ and $\sum_{s=1}^t \mathbb{E}_{s-1} [{\delta'_s}^2] \leq \frac{8 (\sigma^2 + \eta^2/\gamma^2)}{\sigma (1-\sigma)^{2t}S}$.
\end{claim}	

	Now, we are ready to apply Freedman's inequality for martingales.
	
	\begin{theorem}[Freedman's inequality~\citep{freedman1975tail}] \label{thm:freedman}
	Consider a real-valued martingale $\{ Y_k : k \in \mathbb{Z}_{\geq 0} \}$ with respect to some filtration sequence $\{\Sigma_k: k\in \mathbb{Z}_{\geq 0}\}$, and let $\{ X_k : k\in \mathbb{Z}_{>0} \}$ be the associated difference sequence.  Assume that the difference sequence is uniformly bounded: $|X_k| \leq R$ almost surely for $k \in \mathbb{Z}_{>0}$.
	Define the predictable quadratic variation process: $W_k := \sum_{j=1}^k \mathbb{E}_{j-1} \big(X_j^2\big)$
	for $k \in \mathbb{Z}_{>0}$, where $\mathbb{E}_{j-1} \big(\cdot \big) \coloneqq \mathbb{E} \big(\cdot | \Sigma_{j-1}\big)$. Then, for all $t \geq 0$ and $\sigma^2 > 0$,
	$$
	\Pr\left({ \exists k \geq 0 : Y_k \geq t \ \text{ and }\ 
		W_k \leq \sigma^2 }\right)
		\leq \exp \left\{ - \frac{ -t^2/2 }{\sigma^2 + Rt/3} \right\}.
	$$
	\end{theorem}
	
	Applying Freedman's inequality to $\Delta'_{t}$, since the total conditional variance of our martingale difference sequence is bounded almost surely as shown before, we get for any 
	$$r\geq 4 (1-\sigma)^{-t} \left(\sigma+\frac{\eta}{\gamma}\right) \max\left\{ \frac{1}{\sqrt{\sigma S}}\sqrt{\log \frac{2}{\delta}} , \log \frac{2}{\delta}\right\}$$ 
	that
	$$ \Pr(\Delta'_t \geq r) \leq \exp \left\{-\left(\frac{r^2}{2}\right)\bigg/ \left({\frac{8 (\sigma^2 + \eta^2/\gamma^2)}{\sigma (1-\sigma)^{2t}S} + \frac{2(\sigma+\eta/\gamma)r}{(1-\sigma)^t}}\right)\right\} \leq \delta.$$ 
    Hence, using $\Delta'_t = (1-\sigma)^{-t}\Delta_t$, with probability $1-\delta$, we have $$\Delta_t \leq 4  \left(\sigma+\frac{\eta}{\gamma}\right) \left(\frac{1}{\sqrt{{\sigma S}}} \sqrt{\log \frac{2}{\delta}} +\log \frac{2}{\delta}\right).$$ Taking a union bound over the choice of $h$ from $\mathcal{B}\cup \{h^*\}$ and over $t$, along with repeating the argument on the martingale $-\Delta_t$ to furnish the promised two-sided bound, concludes the claim.
\end{proof}

\begin{proof}[Proof of \Cref{cl:phiexp}]
 Using the fundamental theorem of calculus and that $\phi\in \mathcal{C}^2$, which \Cref{lem:phi}.I certifies, we get
	\begin{align*}
		 &\Phi'_\mathcal{D}(H_t, h) \\
		 &= \mathbb{E}_{(x,y)\sim \mathcal{D}}\left[\phi'(yH_{t}(x))yh(x)\right] \\
		 &= \mathbb{E}_{(x,y)\sim \mathcal{D}}\left[\phi'(y(H_{t-1}(x)+\eta h_{t-1}(x)))yh(x)\right] \\
		 &= \mathbb{E}_{(x,y)\sim \mathcal{D}}\left[\phi'(yH_{t-1}(x))yh(x) + \int_0^\eta \phi''(y(H_{t-1}(x)+\eta'h_{t-1}(x))) y^2 h_{t-1}(x) h(x) d\eta' \right]\\
		 &= \mathbb{E}_{(x,y)\sim \mathcal{D}}\left[\phi'(yH_{t-1}(x))yh(x)\right] + \eta \mathbb{E}_{\eta'\sim \text{Unif}[0,\eta]} \mathbb{E}_{(x,y)\sim \mathcal{D}}[h(x) h_{t-1}(x) \phi''(yH_{t-1}^{\eta'}(x))]\\
		 &= \Phi'_\mathcal{D}(H_{t-1}, h) + \eta \mathbb{E}_{\eta'\sim \text{Unif}[0,\eta]}[\Phi''_\mathcal{D}(H_{t-1}^{\eta'}, h, h_{t-1})].
	\end{align*}
	where use the fact that for binary labels $y \in \{-1, 1\}$, $y^2=1$, and the definitions of $\Phi'_\mathcal{D}$ and $\Phi''_\mathcal{D}$. 
\end{proof}

\begin{proof}[Proof of \Cref{cl:dexp1}]
First, we establish a recursive structure on the random random distribution $\mathcal{D}_t$ without any conditional expectation in place.
\begin{claim}\label{cl:dexp2}
For any $t$, we have 
\begin{align*}
		\mathbb{E}_{(x,y)\sim \mathcal{D}_t}& \left[yh(x)\right] = (1-\sigma)\mathbb{E}_{(x,y)\sim \mathcal{D}_{t-1}} \left[yh(x)\right] \\
		&- \frac{\sigma}{\sigma+\eta} \mathbb{E}_{(x,y)\sim \widehat{D}_t}\left[\sigma  \phi'(yH_{t-1}(x))y h(x) + {\eta} \mathbb{E}_{\eta'\sim \text{Unif}[0,\eta]}[\phi''(yH_{t-1}^{\eta'}(x)) h_{t-1}(x)h(x)]\right].
	\end{align*}
\end{claim}

	Using the fact that $\widehat{D}_t$ identically samples data points from $\mathcal{D}$, we arrive at
	\begin{align*}
		& \mathbb{E}_{t-1}[\mathbb{E}_{(x,y)\sim \mathcal{D}_t} \left[yh(x)\right]] \\
		=& (1-\sigma)\mathbb{E}_{(x,y)\sim \mathcal{D}_{t-1}} \left[yh(x)\right] \\
		&- \frac{\sigma^2}{\sigma+\eta} \mathbb{E}_{(x,y)\sim \mathcal{D}}\left[  \phi'(yH_{t-1}(x))y h(x) \right] + \frac{\eta\sigma}{\sigma+\eta} \mathbb{E}_{\eta'\sim \text{Unif}[0,\eta]}\mathbb{E}_{(x,y)\sim\mathcal{D}}\left[\phi''(yH_{t-1}^{\eta'}(x)) h_{t-1}(x)h(x)\right] \\
		=& (1-\sigma)\mathbb{E}_{(x,y)\sim \mathcal{D}_{t-1}} \left[yh(x)\right] - \frac{\sigma^2}{\sigma+\eta} \Phi'_\mathcal{D}(H_{t-1}, h) - \frac{\eta\sigma}{\sigma+\eta} \mathbb{E}_{\eta'\sim \text{Unif}[0,\eta]}[\Phi''_\mathcal{D}(H_{t-1}^{\eta'}, h, h_{t-1})].
	\end{align*}
\end{proof}

\begin{proof}[Proof of \Cref{cl:dexp2}]
By the definition of $\mathcal{D}_t$, we have
	\begin{align*}
		& \mathbb{E}_{(x,y)\sim \mathcal{D}_t} \left[yh(x)\right] \\
		=& (1-\sigma)\mathbb{E}_{(x,y)\sim \mathcal{D}_{t-1}} \left[yh(x)\right] \\
		&- \sigma \mathbb{E}_{(x,y)\sim \widehat{D}_t}\left[h(x) \left( \frac{\sigma}{\sigma+\eta} \phi'(yH_{t-1}(x))y + \frac{\eta}{\eta+\sigma} \mathbb{E}_{\eta'\sim \text{Unif}[0,\eta]}[\phi''(yH_{t-1}^{\eta'}(x)) h_{t-1}(x)] \right)\right]\\
		=& (1-\sigma)\mathbb{E}_{(x,y)\sim \mathcal{D}_{t-1}} \left[yh(x)\right] \\
		&- \frac{\sigma}{\sigma+\eta} \mathbb{E}_{(x,y)\sim \widehat{D}_t}\left[\sigma  \phi'(yH_{t-1}(x))y h(x) + {\eta} \mathbb{E}_{\eta'\sim \text{Unif}[0,\eta]}[\phi''(yH_{t-1}^{\eta'}(x)) h_{t-1}(x)h(x)]\right]
	\end{align*}
	where in the first equality we use the fact that for any binary random variable $Y$ supported on $\{-1, 1\}$ with $\Pr(Y=1) = p$, we have $\mathbb{E}[Y] = 2p-1$.
\end{proof}

\begin{proof}[Proof of \Cref{cl:delta1}]
    Using the recursive expansion of $\Phi'_\mathcal{D}$ (\Cref{cl:phiexp}) and $\mathcal{D}_t$ (\Cref{cl:dexp2}), we have
	\begin{align*}
		\delta'_t =& (1-\sigma)^{-t} \left( \Phi'_\mathcal{D}(H_t, h) - (1-\sigma) \Phi'_\mathcal{D}(H_{t-1}, h) \right)\\
		&+ (1-\sigma)^{-t}\left(1+\frac{\eta}{\sigma}\right) \left(\mathbb{E}_{(x,y)\sim\mathcal{D}_t}[yh(x)] - (1-\sigma)\mathbb{E}_{(x,y)\sim\mathcal{D}_{t-1}}[yh(x)]\right)\\
		=& (1-\sigma)^{-t} \left( \sigma  \Phi'_\mathcal{D}(H_{t-1}, h)+  \eta \mathbb{E}_{\eta'\sim \text{Unif}[0,\eta]}[\Phi''_\mathcal{D}(H_{t-1}^{\eta'}, h, h_{t-1})] \right)\\
		&- (1-\sigma)^{-t} \mathbb{E}_{(x,y)\sim \widehat{D}_t}\left[\sigma  \phi'(yH_{t-1}(x))y h(x) + {\eta} \mathbb{E}_{\eta'\sim \text{Unif}[0,\eta]}[\phi''(yH_{t-1}^{\eta'}(x)) h_{t-1}(x)h(x)]\right]\\
		=& (1-\sigma)^{-t} \sigma \left(  \Phi'_\mathcal{D}(H_{t-1}, h)- \mathbb{E}_{(x,y)\sim \widehat{D}_t}\left[\phi'(yH_{t-1}(x))y h(x)\right]\right) \\
		&+ (1-\sigma)^{-t}\eta \left(\mathbb{E}_{\eta'\sim \text{Unif}[0,\eta]}[\Phi''_\mathcal{D}(H_{t-1}^{\eta'}, h, h_{t-1}) -\mathbb{E}_{(x,y) \sim \widehat{D}_t}[\phi''(yH_{t-1}^{\eta'}(x)) h_{t-1}(x)h(x)] ] \right).
	\end{align*} 
	Using \Cref{lem:phi}, $\phi', \phi''$ are uniformly bounded in magnitude by one, $\Phi'_\mathcal{D}(H, \cdot)$ and $\Phi''_\mathcal{D}(H, \cdot, g)$ are uniformly bounded in magnitude by one for any $H:\mathcal{X}\to\mathbb{R}$ and $1/\gamma$-uniformly bounded function $g$. Hence, ${|\delta'_t|} \leq (1-\sigma)^{-t} 2(\sigma + \eta/\gamma)$. To bound the conditional variance, we use the identity $(a+b)^2 \leq 2a^2 + 2b^2$ in the first line below to show
	\begin{align*}
		&\mathbb{E}_{t-1}[{\delta'_t}^2] \\
		\leq& 2(1-\sigma)^{-2t} \bigg( \sigma^2 \mathbb{E}_{t-1}\left[ \Phi'_\mathcal{D}(H_{t-1}, h)- \mathbb{E}_{(x,y)\sim \widehat{D}_t}\left[\phi'(yH_{t-1}(x))y h(x)\right]\right]^2\\
	 	&+\eta^2 \mathbb{E}_{t-1}\left[\mathbb{E}_{\eta'\sim \text{Unif}[0,\eta]}[\Phi''_\mathcal{D}(H_{t-1}^{\eta'}, h, h_{t-1}) -\mathbb{E}_{(x,y) \sim \widehat{D}_t}[\phi''(yH_{t-1}^{\eta'}(x)) h_{t-1}(x)h(x)] ] \right]^2\bigg)\\
	 	=& 2(1-\sigma)^{-2t}S^{-1} \bigg( \sigma^2 \mathbb{E}_{(x,y)\sim\mathcal{D}}\left[ \Phi'_\mathcal{D}(H_{t-1}, h)- \phi'(yH_{t-1}(x))y h(x)\right]^2\\
	 	&+\eta^2 \mathbb{E}_{(x,y)\sim\mathcal{D}}\left[\mathbb{E}_{\eta'\sim \text{Unif}[0,\eta]}[\Phi''_\mathcal{D}(H_{t-1}^{\eta'}, h, h_{t-1}) -\phi''(yH_{t-1}^{\eta'}(x)) h_{t-1}(x)h(x) ] \right]^2\bigg)\\
	 	\leq & \frac{8 (\sigma^2 + \eta^2/\gamma^2)}{(1-\sigma)^{2t} S},
	\end{align*}
	where we use the fact that the $S$ samples consisting $\widehat{D}_t$ are independent and identically sampled from $\mathcal{D}$ conditioned on $\mathcal{F}_{t-1}$ in the second equality, and that the expectation of any function over $\widehat{D}_t$ is the equivalent sample average. Finally, using an identity on geometric sums, we get 
	$$ \sum_{s=1}^t \mathbb{E}_{s-1} [{\delta'_s}^2] \leq \frac{8 (\sigma^2 + \eta^2/\gamma^2)}{S}\frac{(1-\sigma)^{-2(t+1)}-1}{(1-\sigma)^{-2}-1} \leq \frac{8 (\sigma^2 + \eta^2/\gamma^2)}{\sigma (1-\sigma)^{2t}S}, $$
	where we appeal to the inequality $1-(1-\sigma)^2 = 2\sigma -\sigma^2 \geq \sigma$ for any $\sigma\in [0,1]$.
\end{proof}

\section{Proofs for improved oracle complexity}\label{app:improvp}

\begin{proof}[Proof of \Cref{thm:maincons2}]
The overall proof is similar to that of \Cref{thm:maincons}, with the exception of the following bound on $\varepsilon_{\text{Gen}}$, which we state now (and prove next) in a new lemma relating the empirical correlation on $\mathcal{D}_t$ with $\Phi'_{\mathcal{D}}$. The principal difference between \Cref{lem:vr} and \Cref{lem:vr2} is the presence of a $1/\sqrt{S}$ in both terms on the right in Lemma~\ref{lem:vr2}, however this comes at the cost of a higher polynomial dependence in $\log |\mathcal{B}|$ in the second term on the right side of \Cref{lem:vr2}.
\begin{lemma}\label{lem:vr2}
	There exists a $C>0$ such that with probability $1-\delta$, for all $t\in [T]$ and $h\in \mathcal{B}\cup \{h^*\}$, we have
	$$ \left\lvert \Phi'_\mathcal{D}(H_t, h) + \left(1+\frac{\eta}{\sigma}\right)\corr_{\mathcal{D}_t}(h)\right\rvert\leq \underbrace{C \left(\sigma+\frac{\eta}{\gamma}\right) \left(\frac{1}{\sqrt{{\sigma S}}} \sqrt{\log \frac{|\mathcal{B}|T}{\delta}} +\frac{1}{\sqrt{S}}\left(\log \frac{|\mathcal{B}|T}{\delta}\right)^{3/2}\right)}_{\coloneqq \varepsilon_{\text{Gen}}}. $$
\end{lemma}

From here, with the new definition of $\varepsilon_{\text{Gen}}$ as defined above, identically following the steps in the proof of \Cref{thm:maincons}, which we skip to avoid repetition, we arrive at
\begin{align*}
	 \text{corr}_\mathcal{D}(h^*)-\text{corr}_\mathcal{D}(\overline{h})\leq  \frac{8}{\eta T} + \frac{2\eta}{\gamma^2} + \frac{2\varepsilon_\text{Gen}}{\gamma} +\frac{1}{\gamma}\left(1+\frac{\eta}{\sigma} \right) \left(\varepsilon_0 + \frac{\varepsilon'}{5}\right) + \frac{\varepsilon''}{5}.
\end{align*}
Setting $\varepsilon' = \varepsilon/5\gamma$, $\varepsilon'' = \varepsilon/5$ and plugging in the proposed hyper-parameters with appropriate constants yields the claimed result.
\end{proof}

\begin{proof}[Proof of Lemma~\ref{lem:vr2}.]
The proof of the present lemma is largely similar to that of \Cref{lem:vr}, with two exceptions: (a) we use a variant of Freedman's inequality that applies when the martingale difference sequences (for us, $\delta'_t$ are bounded with high probability, instead of admitting an almost sure absolute bound; (b) to apply this result, we establish a high probability upper bound on $\delta'_t$ that scales as $1/\sqrt{S}$.

Fix any hypothesis $h\in\mathcal{B}$. For any $\eta'\geq 0$, define $H_{t}^{\eta'} = H_t +\eta' h_t$, and 
	$$\Delta_t = \Phi'_\mathcal{D}(H_t, h) + \left(1+\frac{\eta}{\sigma}\right) \mathbb{E}_{(x,y)\sim \mathcal{D}_t} \left[yh(x)\right].$$   
	
As before, adding $(1+\eta/\sigma)$ times the expression in \Cref{cl:dexp1} to the the expansion of $\Phi'_\mathcal{D}(H_t, h)$ in \Cref{cl:phiexp}, we observe that $\mathbb{E}_{t-1}[\Delta_t] = (1-\sigma) \Delta_{t-1}$, and hence conclude that $\Delta'_t = (1-\sigma)^{-t}\Delta_t$ forms a martingale sequence with respect to the $\{\mathcal{F}_t:t\in \mathbb{N}_{\geq 0}\}$ filtration sequence. As shown in \Cref{cl:delta1}, the associated martingale difference sequence $\delta'_t = \Delta'_t - \Delta'_{t-1}$ admits a total variance bound of $\sum_{s=1}^t \mathbb{E}_{s-1} [{\delta'_s}^2] \leq \frac{8 (\sigma^2 + \eta^2/\gamma^2)}{\sigma (1-\sigma)^{2t}S}$.

Now, we state a variant of Freedman's inequality that applies when martingale difference sequences are bounded with high probability.
	
	\begin{theorem}[Freedman's inequality with high probability bounds~\citep{pinelis1994optimum,371436}] \label{thm:freedman2}
	Fix a positive integer $n$. Consider a real-valued martingale $\{ Y_k : k \in \mathbb{Z}_{\geq 0} \}$ with respect to some filtration sequence $\{\Sigma_k: k\in \mathbb{Z}_{\geq 0}\}$, and let $\{ X_k : k\in \mathbb{Z}_{>0} \}$ be the associated difference sequence.  Assume that the difference sequence is uniformly bounded with high probability for some $R,\delta'$:
    $$\Pr\left(\max_{k\in [n]} |X_k| \geq R\right) \leq \delta'. $$ 
	Define the predictable quadratic variation process: $W_n := \sum_{j=1}^n \mathbb{E}_{j-1} \big(X_j^2\big)$, where $\mathbb{E}_{j-1} \big(\cdot \big) \coloneqq \mathbb{E} \big(\cdot | \Sigma_{j-1}\big)$. Then, for all $t \geq 0$ and $\sigma^2 > 0$,
	$$
	\Pr\left({ Y_n \geq t \ \text{ and }\ 
		W_n \leq \sigma^2 }\right)
		\leq \exp \left\{ - \frac{ -t^2/2 }{\sigma^2 + Rt/3} \right\} +\delta'.
	$$
	\end{theorem}
	
To apply this variant of Freedman's inequality, we establish a high probability bound on $\Delta'_{t}$, that is, ignoring logarithmic terms, substantially better than the almost sure bound in \Cref{cl:delta1}.

\begin{claim}\label{cl:delta2}
There exists a universal constant $C$, such that for any $t$, we have$$\Pr\left(\max_{s\in [t]} |\delta'_s| \geq \frac{C}{(1-\sigma)^t}\left(\sigma+\frac{\eta}{\gamma}\right)\frac{1}{\sqrt{S}}\sqrt{\log \frac{t}{\delta}}\right)\leq \delta.$$
\end{claim}

Combining this with the almost sure bound on the total conditional variance of our martingale difference sequence, we get for any 
	$$r\geq 4 (1-\sigma)^{-t} \left(\sigma+\frac{\eta}{\gamma}\right) \max\left\{ \frac{1}{\sqrt{\sigma S}}\sqrt{\log \frac{2}{\delta}} , \frac{C}{\sqrt{S}}\left(\log \frac{2t}{\delta}\right)^{3/2}\right\}$$ 
	that
	$$ \Pr(\Delta'_t \geq r) \leq \exp \left\{-\left(\frac{r^2}{2}\right)\bigg/ \left({\frac{8 (\sigma^2 + \eta^2/\gamma^2)}{\sigma (1-\sigma)^{2t}S} + \frac{C(\sigma+\eta/\gamma)r}{(1-\sigma)^t\sqrt{S}}} \sqrt{\log\frac{t}{\delta}}\right)\right\} +\delta \leq 2\delta.$$ 
    Hence, using $\Delta'_t = (1-\sigma)^{-t}\Delta_t$, with probability $1-2\delta$, we have $$\Delta_t \leq 4  \left(\sigma+\frac{\eta}{\gamma}\right) \left(\frac{1}{\sqrt{{\sigma S}}} \sqrt{\log \frac{2}{\delta}} +\frac{C}{\sqrt{S}}\left(\log \frac{2t}{\delta}\right)^{3/2}\right).$$ Taking a union bound over the choice of $h$ from $\mathcal{B}\cup \{h^*\}$ and over $t$, along with repeating the argument on the martingale $-\Delta_t$ to furnish the promised two-sided bound, concludes the claim.
\end{proof}

\begin{proof}[Proof of \Cref{cl:delta2}]
The proof begins similarly to the one for \Cref{cl:delta1}. Using the recursive expansion of $\Phi'_\mathcal{D}$ (\Cref{cl:phiexp}) and $\mathcal{D}_t$ (\Cref{cl:dexp2}), we have
	\begin{align*}
		\delta'_t =& (1-\sigma)^{-t} \left( \Phi'_\mathcal{D}(H_t, h) - (1-\sigma) \Phi'_\mathcal{D}(H_{t-1}, h) \right)\\
		&+ (1-\sigma)^{-t}\left(1+\frac{\eta}{\sigma}\right) \left(\mathbb{E}_{(x,y)\sim\mathcal{D}_t}[yh(x)] - (1-\sigma)\mathbb{E}_{(x,y)\sim\mathcal{D}_{t-1}}[yh(x)]\right)\\
		=& (1-\sigma)^{-t} \left( \sigma  \Phi'_\mathcal{D}(H_{t-1}, h)+  \eta \mathbb{E}_{\eta'\sim \text{Unif}[0,\eta]}[\Phi''_\mathcal{D}(H_{t-1}^{\eta'}, h, h_{t-1})] \right)\\
		&- (1-\sigma)^{-t} \mathbb{E}_{(x,y)\sim \widehat{D}_t}\left[\sigma  \phi'(yH_{t-1}(x))y h(x) + {\eta} \mathbb{E}_{\eta'\sim \text{Unif}[0,\eta]}[\phi''(yH_{t-1}^{\eta'}(x)) h_{t-1}(x)h(x)]\right]\\
		=& (1-\sigma)^{-t} \sigma \left( \underbrace{\Phi'_\mathcal{D}(H_{t-1}, h)- \mathbb{E}_{(x,y)\sim \widehat{D}_t}\left[\phi'(yH_{t-1}(x))y h(x)\right]}_{\coloneqq A_1}\right) \\
		&+ (1-\sigma)^{-t}\eta \left(\mathbb{E}_{\eta'\sim \text{Unif}[0,\eta]}[\underbrace{\Phi''_\mathcal{D}(H_{t-1}^{\eta'}, h, h_{t-1}) -\mathbb{E}_{(x,y) \sim \widehat{D}_t}[\phi''(yH_{t-1}^{\eta'}(x)) h_{t-1}(x)h(x)}_{\coloneqq A_2}] ] \right).
	\end{align*} 
    However, this time, we note that, conditioned on $\mathcal{F}_{t-1}$, $A_1$ and $A_2$ are averages of $S$ zero-mean IID random variables, where each constituent is absolutely bounded by one in magnitude (\Cref{lem:phi}). Hence, by Hoeffding's inequality, with $1-2\delta$, we have that 
    $$ \max\{|A_1|, |A_2|\} \leq \frac{100}{\sqrt{S}}\sqrt{\log \frac{2}{\delta}}.$$
    Note that since this statement holds true for all realizations in $\mathcal{F}_{t-1}$ -- in words, it is a statement about the randomness in the present round alone -- it remains true when not subject to such filtration's, i.e., by marginalizing over $\mathcal{F}_{t-1}$. A union bound over $t$ now yields the claim
\end{proof}

\section{Proofs for extensions to infinite classes}\label{app:inf}
\begin{definition}[Covering number]
Given a set $\mathcal{F}$ in linear space $\mathcal{V}$, a semi norm $\|\cdot\|$ on $\mathcal{V}$, $\mathcal{N}(\varepsilon, \mathcal{F}, \|\cdot \|)$ is the size of the smallest set $\mathcal{G}$ such that for all $f\in \mathcal{F}$, there exists a $g\in \mathcal{G}$ such that $\|f-g\|\leq \varepsilon$.
\end{definition}

\begin{proof}[Proof of Theorem~\ref{thm:mainvc}]
First, we state (and subsequently prove) the following sample complexity result using $L_1$ covering numbers. We note that it may be possible to improve this result for specific classes of functions, i.e., monotonic functions, by applying chaining techniques \citep{dudley1974metric} to $L_2$ distances.

\begin{theorem}[Main result for Covering Number]\label{thm:maincover}
Define $L_{1,\mathcal{D}_\mathcal{X}}(f,g) = \mathbb{E}_{x\sim \mathcal{D}_\mathcal{X}}[|f(x)-g(x)|]$, for any two functions $f,g:\mathcal{X}\to \mathbb{R}$, where $\mathcal{D}_\mathcal{X}$ is the marginal distribution of $\mathcal{D}$ on the features set $\mathcal{X}$. There exists a setting of parameters such that for any for any $\gamma$-agnostic weak learning oracle with fixed tolerance $\varepsilon_0$ and failure probability $\delta_0$, Algorithm~\ref{alg:rev} 
produces a hypothesis $\overline{h}$ such that with probability $1-10\delta_0 T-10\delta T$,
	$$ \corr_\mathcal{D}(\overline{h}) \geq \max_{h\in \mathcal{H}}\corr_\mathcal{D}(h) - \frac{2\varepsilon_0}{\gamma} - \varepsilon, $$
while making $T= \mathcal{O}((\log \mathcal{N}(\varepsilon/10\gamma, \mathcal{B}, L_{1,\mathcal{D}_\mathcal{X}}))/\gamma^2\varepsilon^2)$ calls to the weak learning oracle, and sampling $TS+S_0=\mathcal{O}((\log \mathcal{N}(\varepsilon/10\gamma, \mathcal{B}, L_{1,\mathcal{D}_\mathcal{X}}))/\gamma^3\varepsilon^3)$ labeled examples from $\mathcal{D}$.
\end{theorem}

The claim follows immediately by the application of the following result from \cite{van1997weak} to place an upper bound of the $L_1$ covering in terms of the VC dimension in \Cref{thm:maincover}. This result originally due to \cite{haussler1995sphere} was established for distances defined by $n$-point empirical measures, for some finite $n$. The version in \cite{van1997weak} works on arbitrary distributions, by first proving that by a result proven for empirical-type measures transfers without loss to any distribution.
\begin{theorem}[Theorem 2.6.4 in \cite{van1997weak}]
There exists a universal constant $C$ such that for any VC class $\mathcal{B}$, any probability measure $\mathcal{D}_\mathcal{X}$, and $0\leq \varepsilon\leq 1$, $$\mathcal{N}(\varepsilon, \mathcal{B}, L_{1,\mathcal{D}_\mathcal{X}}) \leq C \cdot \text{VC}(\mathcal{B}) \cdot\left(\frac{4e}{\varepsilon}\right)^{\text{VC}(\mathcal{B})}.$$
\end{theorem}
This concludes the proof.
\end{proof}

\begin{proof}[Proof of Theorem~\ref{thm:maincover}]
We first invoke \Cref{lem:vr} but on a minimal $\varepsilon_{\text{Cov}}$-cover (say $\mathcal{B}_\text{Cov}$) of $\mathcal{B}$ with respect to the $L_{1,\mathcal{D}_\mathcal{X}}$ semi norm to get that there exists a $C>0$ such that with probability $1-\delta$, for all $t\in [T]$ and $h\in \mathcal{B}_\text{Cov}\cup \{h^*\}$, we have
\begin{align*} 
&\left\lvert \Phi'_\mathcal{D}(H_t, h) + \left(1+\frac{\eta}{\sigma}\right)\text{corr}_{\mathcal{D}_t}(h)\right\rvert\\
&\leq \underbrace{C\left(\sigma + \frac{\eta}{\gamma}\right)\left(\frac{1}{\sqrt{\sigma S}}\sqrt{\log\frac{T\log \mathcal{N}(\varepsilon_{\text{Cov}}, \mathcal{B}, L_{1,\mathcal{D}_\mathcal{X}})}{\delta}} + \log \frac{T\log \mathcal{N}(\varepsilon_\text{Cov}, \mathcal{B}, L_{1,\mathcal{D}_\mathcal{X}})}{\delta}\right)}_{\coloneqq \varepsilon'_{\text{Gen}}}.
\end{align*}
Fix any $g\in \mathcal{B}$. Now, by virtue of the $\varepsilon_{\text{Cov}}$-cover, there exists a $h\in \mathcal{B}$ such that $L_{1,\mathcal{D}_\mathcal{X}}=\mathbb{E}_{x\sim \mathcal{D}_\mathcal{X}}[|f(x)-g(x)|]\leq \varepsilon_{\text{Cov}}$. Now, using the definition of $\Phi'_\mathcal{D}$ and a uniform (absolute) upper bound on $\phi'$ (\Cref{lem:phi}.III), we get
\begin{align*}
    |\Phi'_\mathcal{D}(H_t, h)-\Phi'_\mathcal{D}(H_t, g)| &= |\mathbb{E}_{(x,y)\sim \mathcal{D}}[y\phi'(yH_t(x)) (h(x)-g(x))]|\\
    &\leq \mathbb{E}_{(x,y)\sim \mathcal{D}}[|y\phi'(yH_t(x))| \cdot |h(x)-g(x)|] \\
    &\leq \mathbb{E}_{(x,y)\sim \mathcal{D}}[ |h(x)-g(x)|] \\
    &\leq \mathbb{E}_{x\sim \mathcal{D}_\mathcal{X}}[ |h(x)-g(x)|] \leq \varepsilon_{\text{Cov}}.
\end{align*}
Similarly, using the fact that the marginal distribution of $\mathcal{D}_t$ on $\mathcal{X}$ is the same as $\mathcal{D}_\mathcal{X}$, we get
\begin{align*}
    |\text{corr}_{\mathcal{D}_t}(h)-\text{corr}_{\mathcal{D}_t}(g)| \leq \mathbb{E}_{x\sim \mathcal{D}_\mathcal{X}}[ |h(x)-g(x)|] \leq \varepsilon_{\text{Cov}}.
\end{align*}

Combining these, we have that with probability $1-\delta$, for all $t\in [T]$ and $h\in \mathcal{B}_\text{Cov}\cup \{h^*\}$, we have
$$
\left\lvert \Phi'_\mathcal{D}(H_t, h) + \left(1+\frac{\eta}{\sigma}\right)\text{corr}_{\mathcal{D}_t}(h)\right\rvert \leq { \varepsilon'_{\text{Gen}}} + 2\varepsilon_{\text{Cov}}. $$
From here, proceeding identically as the steps in the proof of \Cref{thm:maincons} yields
\begin{align*}
	 \text{corr}_\mathcal{D}(h^*)-\text{corr}_\mathcal{D}(\overline{h})\leq  \frac{8}{\eta T} + \frac{2\eta}{\gamma^2} + \frac{2(\varepsilon'_\text{Gen}+\varepsilon_{\text{Cov}})}{\gamma} +\frac{1}{\gamma}\left(1+\frac{\eta}{\sigma} \right) \left(\varepsilon_0 + \frac{\varepsilon'}{5}\right) + \frac{\varepsilon''}{5}.
\end{align*}
Setting $\varepsilon' = \varepsilon/5\gamma$, $\varepsilon'' = \varepsilon/5$, $\varepsilon_{\text{Cov}}=\varepsilon/10\gamma$ and plugging in the proposed hyper-parameters with appropriate constants yields the claimed result.
\end{proof}

\section{Boosting for reinforcement learning}\label{app:rl}
{\bf MDP.} In this section, we consider a Markov Decision Process $\mathcal{M}=(\mathcal{S}, \mathcal{A}, r, P, \beta, \mu_0)$, where $\mathcal{S}$ is a set of states, $\mathcal{A}=\{\pm 1\}$ is a binary set of actions, $r:\mathcal{S}\times \mathcal{A}\to [0,1]$ determines the (expected) reward at any state-action pair, $P:\mathcal{S}\times \mathcal{A}\to \mathcal{S}$ captures the transition dynamics of the MDP, i.e., $P(s'|s,a)$ is the probability of moving to state $s'$ upon taking action $a$ at state $s$, $\beta\in [0,1)$ is the discount factor, and $\mu_0$ is the initial state distribution. Without any loss, one may restrict their consideration to Markovian policies \citep{puterman2014markov} of the form $\pi:\mathcal{S}\to \Delta(\mathcal{A})$, where an agent at each point in time chooses action $a$ at state $s$ independently with probability $\pi(a|s)$.

For any state $s\in \mathcal{S}$, action $a\in \mathcal{A}$, and distribution $\mu \sim \Delta(\mathcal{S})$ over states, define  state-action and state-value functions as 
\begin{align*}
    Q^\pi(s,a) &= \mathbb{E}\left[\sum_{t=0}^\infty \beta^t r(s_t,a_t) \bigg| \pi, s_0=s, a_0=a\right], \\
    V^\pi(s) &= \mathbb{E}\left[\sum_{t=0}^\infty \beta^t r(s_t,a_t) \bigg| \pi, s_0=s\right], \\
    V^\pi_\mu &= \mathbb{E}_{s\sim \mu}\left[V^\pi(s)\right]. \\
\end{align*}
We will abbreviate $V^\pi_{\mu_0}=V^\pi$, since it captures the value of any policy starting from the canonical starting state distribution. Finally, the occupancy measure $\mu^\pi_{\mu'}$ induced by a policy $\pi$ starting from an initial state distribution $\mu'$ is stated below. We will take $\mu^\pi = \mu^\pi_{\mu_0}$ as a matter of convention.

{\bf Accessing the MDP.} Following \cite{brukhim2022boosting}, we consider two models of accessing the MDP; we will furnish a different result for each. In the {\em episodic model}, the learner interacts with the MDP in a limited number of episodes of reasonable length (i.e., $\approx (1-\beta)^{-1}$), and the starting state of MDP is always drawn from $\mu_0$. In the second, termed {\em rollouts with $\nu$-resets}, the learner's interaction is still limited to a small number of episodes, however, the MDP now samples its starting state from $\nu$. It is important to stress that in both cases, the learner's objective is the same, to maximize $V^\pi$ starting from $\mu_0$. However, $\nu$ could be more {\em spread out} over the state space than $\mu_0$, and provide an implicit source of explanation, and the learner's guarantee as shown next benefits from its dependence on a milder notion of distribution mismatch in this case.

{\bf Weak Leaner.} For convenience, we restate our weak learning definition, this time using $\pi$ to denote policies, instead of $h$. Note that our definition is equivalent to that used by \cite{brukhim2022boosting}, because for binary actions a random policy induces an accuracy of half regardless of the distribution over features and labels. One may use the identity $\text{corr}_\mathcal{D}(\pi) = 1-2l_{\mathcal{D}}(\pi)$ to observe this. In fact, our assumption of weak learning might ostensibly seem weaker, since it operates with $0/1$ losses (equivalently, correlations), whereas the losses in previous work are assumed general linear. However, for binary actions, this difference is insubstantial and purely stylistic.

\begin{definition}[Agnostic Weak Learner]\label{def:wlrl}
A learning algorithm is called a $\gamma$-agnostic weak learner with sample complexity $m:\mathbb{R}_+\times \mathbb{R}_+ \to \mathbb{N}\cup \{+\infty\}$ with respect to a policy class $\Pi$ and a base policy class $\mathcal{B}$ if, for any $\varepsilon_0, \delta_0>0$, upon being given $m(\varepsilon_0, \delta_0)$ independently and identically distributed samples from any distribution $\mathcal{D}'\in \Delta(\mathcal{X}\times \{\pm 1\})$, it can output a base policy $\mathcal{W}\in \mathcal{B}$ such that with probability $1-\delta_0$ we have
$$ \text{corr}_\mathcal{D'}(\mathcal{W}) \geq \gamma \max_{\pi\in \Pi}\text{corr}_\mathcal{D'}(\pi) -\varepsilon_0.$$
\end{definition} 

{\bf Policy Completeness and Distribution Mismatch.}  $\pi^*\in\argmax_{\pi} V^\pi$ be a reward maximizing policy, and $V^*$ be its value. Let $\mathbbl{\Pi}$ be the convex hull of the boosted policy class, i.e., the outputs of the boosting algorithm. For any state distribution $\mu'$, define the policy completeness $\mathcal{E}_{\mu'}$ term as 
$$ \mathcal{E}_{\mu'} = \max_{\pi\in \mathbbl{\Pi}} \min_{\pi'\in \Pi} \mathbb{E}_{s\sim \mu^\pi_{\mu'}} [\max_{a\in \mathcal{A}} Q^\pi(s,a) - \mathbb{E}_{a\sim \pi'(\cdot|s)}Q^\pi(s,a)].$$
In words, this term captures how well the greedy policy improvement operator is approximated by $\Pi$ in an state-averaged sense over the  distribution induces by any policy in $\mathbbl{\Pi}$. Finally, we define distribution mismatch coefficients below.
$$ C_\infty = \max_{\pi\in \mathbbl{\Pi}} \|\mu^{\pi^*}/\mu^\pi\|_\infty, \quad D_\infty = \|\mu^{\pi^*}/\nu\|_\infty. $$

\begin{algorithm}[t]
    \caption{RL Boosting adapted from \cite{brukhim2022boosting}}\label{alg:rlMAIN1}
    \begin{algorithmic}[1]
        \STATE \textbf{Input}: iteration budget $T$, state distribution $\mu$, step sizes $\eta_{t}$, post-selection sample size $P$
        \STATE Initialize a policy $\pi_0\in \Pi$ arbitrarily. 
        \FOR{$t=1$ {\bfseries to} $T$}
        \STATE Run \Cref{alg:rev} to get $\pi'_t$, while using \Cref{alg:q_sampler} to produce a distribution over state-actions (ignore $\widehat{Q}$) by executing the current policy $\pi_{t-1}$ starting from the initial state distribution $\mu$.
        \STATE Update $\pi_t = (1-\eta_{t})\pi_{t-1} + \eta_{t} \pi'_t$.
        \ENDFOR
    \STATE Run each policy $\pi_t$ for $P$ rollouts to compute an empirical estimate $\widehat{V^{\pi_t}}$  of the expected return.
    \RETURN $\overline{\pi} = \pi_{t'}$ where $t'=\argmax_t \widehat{V^{\pi_t}}$.
    \end{algorithmic}
\end{algorithm}

\begin{algorithm}[t]
    \begin{algorithmic}[1]
        \STATE Sample state $s_0 \sim \mu$ and action  $a' \sim \text{Unif}(\mathcal{A})$.
        \STATE Sample $s\sim \mu^{\pi}$ as follows:
        at every step $h$,  with probability $\beta$, execute $\pi$; else, accept  $s_h$. 
        \STATE\label{state:next_step} Take action $a'$ at state $s_h$, then
        continue to execute $\pi$, and use a termination probability of
        $1-\beta$. Upon termination, set
        $R(s_h,a')$ as the sum
        of rewards from time $h$ onwards.
        \STATE Define the vector $\widehat{Q}$, such that for all $a \in A$, $\widehat{Q}(a) = 2R(s_h,a') \cdot \mathbb{I}_{a=a'}$.
        \STATE With probability $C\widehat{Q}(a')$, set $y=a'$ else set $y\in \mathcal{A}-\{a'\}$, where $C=(1-\beta)/2$.
        \RETURN $(s_h, \widehat{Q}, y)$.
    \end{algorithmic}
    \caption{Trajectory Sampler adapted from \cite{brukhim2022boosting}}
    \label{alg:q_sampler}
\end{algorithm}

\begin{theorem}\label{thm:rl2}
Let $\mathcal{W}$ be a $\gamma$-weak learner for the policy class $\Pi$ operating with a base class $\mathcal{B}$, with sample complexity $m(\varepsilon_0, \delta_0) = (\log |\mathcal{B}|/\delta_0)/\varepsilon_0^2$. Fix tolerance $\varepsilon$ and failure probability $\delta$. In the {\em episodic} access model, there is a setting of parameters such that \Cref{alg:rlMAIN1} when given access to $\mathcal{W}$ produces a policy $\overline{\pi}$ such that with probability $1-\delta$, we have
$$ V^* - V^{\overline{\pi}} \leq \frac{C_\infty \mathcal{E}}{1-\beta} + \varepsilon,$$
while sampling $$\mathcal{O}\left(\frac{C_\infty^5\log |\mathcal{B}|}{(1-\beta)^9 \gamma^3 \varepsilon^4}\right)$$ episodes of length $\mathcal{O}((1-\beta)^{-1})$. In the {\em $\nu$-reset} access model, there is a setting of parameters such that \Cref{alg:rlMAIN1} when given access to $\mathcal{W}$ produces a policy $\overline{\pi}$ such that with probability $1-\delta$, we have
$$ V^* - V^{\overline{\pi}} \leq \frac{D_\infty \mathcal{E}_\nu}{(1-\beta)^2} + \varepsilon,$$
while sampling $$\mathcal{O}\left(\frac{D_\infty^5\log |\mathcal{B}|}{(1-\beta)^{15}\gamma^3 \varepsilon^5}\right)$$ episodes of length $\mathcal{O}((1-\beta)^{-1})$. 
\end{theorem}

\section{Proofs for boosting for reinforcement learning}\label{app:rl2}
\begin{proof}[Proof of \Cref{thm:rl2}]
The proof here closely follows that of Theorem 7 in \cite{brukhim2022boosting}, and we only indicate the necessary departures. Since we utilize the outer algorithm from previous work, the associated guarantees naturally carry over. The departure comes from our substitution of the {\em internal boosting} procedure in \Cref{alg:rlMAIN1} with \Cref{alg:rev}; in fact, in its place \cite{brukhim2022boosting} use a result of \cite{hazan2021boosting} which can be seen as a generalization of \cite{kanade2009potential}, e.g., it uses fresh samples for every round of boosting, to other non--binary action sets preserving its $\approx 1/\varepsilon^4$ sample complexity. In this view, the improvement in sample complexity by using the present paper's apprach seems natural.

For the episodic model, applying the second part of Theorem 9 in \cite{brukhim2022boosting}, while noting the smoothness of $V^\pi$, and combining the result with Lemma 18 and Lemma 11 in \cite{brukhim2022boosting}, we have with probability $1-T\delta$
    \begin{align*}
        V^*-V^{\bar{\pi}} \leq & \frac{C_\infty\mathcal{E}}{1-\beta} +\frac{4C^2_\infty}{(1-\beta)^3 T} + \frac{C_\infty}{1-\beta} \left(\max_{\pi\in\Pi, t\in [T]}\mathbb{E}_{(s,\widehat{Q}, y)\sim \mathcal{D}_t} [ \widehat{Q}^\top (\pi(\cdot|s)-\pi'_t(\cdot|s))]\right),
    \end{align*}
where $\pi'_t$ is the output of the internal boosting algorithm in Line 4 of \Cref{alg:rlMAIN1}, and $\mathcal{D}_t$ is the distribution produced by \Cref{alg:q_sampler} with $\pi_{t-1}$ selected as the policy for execution. Here, by explicit calculation, noting $a'\sim \text{Unif}(\mathcal{A})$, one may verify  for any $t$ that
$$ \mathbb{E}_{(s,\widehat{Q},y)\sim \mathcal{D}_t} [y|s] = \frac{1-\beta}{2} \mathbb{E}_{(s,\widehat{Q},y)\sim \mathcal{D}_t} [\widehat{Q}(+1)-\widehat{Q}(-1)|s].$$
Recall that $\mathcal{A}=\{\pm 1\}$, and hence $\pi:\mathcal{S}\to \Delta(\{\pm 1\})$. Hence, for any policy pair $\pi_1, \pi_2$ we have 
$$(1-\beta)\mathbb{E}_{(s,\widehat{Q}, y)\sim \mathcal{D}_t} [ \widehat{Q}^\top (\pi_1(\cdot|s)-\pi_2(\cdot|s))] = \mathbb{E}_{(s,\widehat{Q},y)\sim \mathcal{D}_t} \mathbb{E}_{a_1\sim \pi_1(s)}\mathbb{E}_{a_2\sim \pi_2(s)}[y(a_1-a_2)].$$
Therefore, all we need to ensure is that output of \Cref{alg:rev} as instantiated in \Cref{alg:rlMAIN1} every round has an excess correlation gap over the best policy $\Pi$ no more that $(1-\beta)^2\varepsilon/C_\infty$, which \Cref{alg:rev} assures us can be accomplished with $\mathcal{O}\left(\frac{C_\infty^3 \log |\mathcal{B}|}{(1-\beta)^6\gamma^3 \varepsilon^3}\right)$ samples. The total number of samples is $T=\mathcal{O}\left(\frac{C_\infty^2}{(1-\beta)^3 \varepsilon}\right)$ times greater.

Similarly, for the $\nu$-reset model, applying the first part of Theorem 9 in \cite{brukhim2022boosting}, and combining the result with Lemma 19 and Lemma 11 in \cite{brukhim2022boosting}, we have 
    \begin{align*}
        V^*-V^{\bar{\pi}} \leq \frac{D_\infty\mathcal{E}_\nu}{(1-\beta)^2} &+ \frac{2D_\infty }{(1-\beta)^3 \sqrt{T}} + \frac{D_\infty}{(1-\beta)^2} \left(\max_{\pi\in\Pi, t\in [T]}\mathbb{E}_{(s,\widehat{Q}, y)\sim \mathcal{D}_t} [ \widehat{Q}^\top (\pi(\cdot|s)-\pi'_t(\cdot|s))]\right)\\ &+ \frac{96 D_\infty}{(1-\beta)^3\sqrt{P}}\log \frac{1}{\delta}
    \end{align*}
    Again, we need to ensure is that output of \Cref{alg:rev} as instantiated in \Cref{alg:rlMAIN1} every round has an excess correlation gap over the best policy $\Pi$ no more that $(1-\beta)^3\varepsilon/D_\infty$, which \Cref{alg:rev} assures us can be accomplished with $\mathcal{O}\left(\frac{D_\infty^3 \log |\mathcal{B}|}{(1-\beta)^9\gamma^3 \varepsilon^3}\right)$ samples. The total number of samples is $T=\mathcal{O}\left(\frac{D_\infty^2}{(1-\beta)^6 \varepsilon^2}\right)$ times greater.
\end{proof}

\section{Proofs for agnostic learning of halfspaces}\label{app:half}
\begin{proof}[Proof of \Cref{thm:half}]
Using an approximation result of \cite{klivans2004learning}, we observe that ERM on the Fourier basis $\chi_S(x) = \prod_{i \in S} x_i$, namely parities on subsets $S$, can be used to produce a weak learner. 
This result guarantees that an $n$-dimensional halfspace can be approximated with uniform weighting on the hypercube
to $\varepsilon^2$ $\ell_2$-error using degree-limited $\mathcal{B}_{n,d} = \{\pm \chi_S: |S| \le d \}$ as a basis, where $d = 20 \varepsilon^{-4}$.
As a result, at least one $h \in \mathcal{B}_{n,d}$ must have high correlation. 

\begin{lemma}[Lemma 5 in \cite{kalai2008agnostic}]\label{lem:half_weak}
    Let $\mathcal{D}$ be any data distribution over $ \{\pm 1\}^n \times \{-1, 1\}$ with marginal distribution $\text{Unif}(\{\pm 1\}^n)$ on the features. For any fixed $\varepsilon$ and $d = 20 \varepsilon^{-4}$, there exists some $h \in \mathcal{B}_{n,d}$ such
    \begin{align*}
        \corr_{\mathcal{D}}(h) \geq \frac{\max_{c \in \mathcal{H}} \corr_{\mathcal{D}}(c) - \varepsilon}{n^d}
    \end{align*}
\end{lemma}

The result follows directly from the preceding lemma, which provides a weak learner for the task, and \Cref{thm:maincons}. We note that $|\mathcal{B}_{n,d}| < n^d$ and $\gamma = n^{-d}$, so 
    $$ \frac{\log |\mathcal{B}_{n,d}|}{\gamma^3\varepsilon^3} \leq \frac{dn^{3d}\log(n)}{\varepsilon^3}. $$
\end{proof}

\section{Additional experimental details}\label{sec:exp_app}
For all algorithms, we perform a grid search on the number of boosting rounds with $T \in \{25, 50, 100\}$. For \Cref{alg:rev} we search over $\sigma \in \{0.1, 0.25, 0.5\}$ as well. Rather than using a fixed $\eta$, our implementation uses an adaptive step-size scheme proportional to the empirical correlation on the current relabeled distribution. Our experiments were performed using the fractional relabeling scheme stated in \cite{kanade2009potential}, intended to reduce the stochasticity the algorithm is subject to. In particular, rather than sampling labels, we provide both $(x, y)$ and $(x, -y)$ in our dataset with weights $\frac{1 + w^t(x, y)}{2}$ and $\frac{1 - w^t(x, y)}{2}$ respectively. For the baselines,  $w^t(x, y) = - \phi'_{\text{mad}}(H_{t}y) = \min(1, \exp(- H_{t}(x)y))$, where $\phi_{\text{mad}}$ is the Madaboost potential \citep{domingo2000madaboost}. For the implementation of the proposed algorithm, for greater reproducibility, we use the weighting function $w^t(x,y) = (1-\sigma)\phi'(H_{t-1}(x)y) - \phi'(H_{t}(x)y)$, which is analytically equivalent to computing the expectation of $p_t(x,y,\eta')$ in \Cref{alg:rev} over $\eta'\sim \text{Unif}[0,\eta]$. The runtime used for all experiments was a Google Cloud Engine VM instance with 2 vCPUs (Intel Xeon 64-bit @ 2.20 GHz) and 12 GB RAM.

\section{Guide for practical adaptation of \Cref{alg:rev}}\label{sec:guide_app}
For many hyperparamters, our theory provides strong clues (the link between $\eta$ and $\sigma$). Briefly, in practice, given a fixed dataset with $m_{\text{total}}$ data points, there are two parameters we think a practitioner should concern herself with: the mixing parameter $\sigma$ and the number of rounds of boosting $T$, while the rest can be determined, or at least guessed well, given these. The first choice $\sigma$ dictates the relative weight ascribed to reused samples across rounds of boosting, while $T$, apart from determining the running time and generalization error, implicitly limits the algorithm to using $m_{\text{total}}/T$ fresh samples each round. For $\eta$, one may use the adaptive step-size schedule suggested in the previous section, which also obviates the difficulty of selecting $\gamma$. Similarly, our branching criteria, whose necessity is explained in \Cref{app:error}, although theoretically both sound and necesssary, is overly conservative. In practice, we believe choosing the better of $-\text{sign}(H_t)$ and $W_t$ whichever produces the greatest empirical distribution on the relabeled distribution will perform best.


\end{document}